\documentclass[sigconf]{acmart}
\usepackage{amsmath,amsfonts}
\usepackage{algorithmic}
\usepackage[noline,ruled,algo2e,linesnumbered]{algorithm2e}
\usepackage{comment}
\usepackage{multirow}
\usepackage{hyperref}
\usepackage{graphicx}
\usepackage{xcolor}
\usepackage{subcaption}
\usepackage[normalem]{ulem}

\AtBeginDocument{%
  \providecommand\BibTeX{{%
    \normalfont B\kern-0.5em{\scshape i\kern-0.25em b}\kern-0.8em\TeX}}}

\usepackage[colorinlistoftodos]{todonotes}

\newcommand{\romain}[1]{\todo[size=\small,inline,color=purple!50]{#1  \-- Romain}}

\newcommand{\edit}[1]{{\color{black}#1}}

\DeclareMathOperator*{\argmax}{arg\,max}
\DeclareMathOperator{\st}{s.t.}
\DeclareMathOperator*{\argmin}{arg\,min}
\newcommand{\ts}{\textsuperscript}

\begin{document}

\copyrightyear{2021} 
\acmYear{2021} 
\setcopyright{licensedusgovmixed}\acmConference[SC '21]{The International Conference for High Performance Computing, Networking, Storage and Analysis}{November 14--19, 2021}{St. Louis, MO, USA}
\acmBooktitle{The International Conference for High Performance Computing, Networking, Storage and Analysis (SC '21), November 14--19, 2021, St. Louis, MO, USA}
\acmPrice{15.00}
\acmDOI{10.1145/3458817.3476203}
\acmISBN{978-1-4503-8442-1/21/11}

\title{AgEBO-Tabular: Joint Neural Architecture and Hyperparameter Search with Autotuned Data-Parallel Training for Tabular Data}

\author{Romain \'Egel\'e}
\affiliation{%
  \institution{\'Ecole polytechnique}
%   \city{Pa}
   \country{Palaiseau, France}
}
\email{romain.egele@polytechnique.edu}

\author{Prasanna Balaprakash}
\affiliation{%
  \institution{Argonne National Laboratory}
%   \city{Chicago}
   \country{Lemont, Illinois, USA}
  }
\email{pbalapra@anl.gov}

\author{Isabelle Guyon}
\affiliation{%
  \institution{CNRS/Inria-LISN, U. Paris-Saclay}
   \country{France}
}
\email{guyon@chalearn.org}

\author{Venkatram Vishwanath}
\affiliation{%
 \institution{Argonne National Laboratory}
   \country{Lemont, Illinois, USA}
}
\email{venkat@anl.gov}

\author{Fangfang Xia}
\affiliation{%
  \institution{Argonne National Laboratory}
   \country{Lemont, Illinois, USA}
}
\email{fangfang@anl.gov}

\author{Rick Stevens}
\affiliation{%
  \institution{Argonne National Laboratory}
   \country{Lemont, Illinois, USA}
}
\email{stevens@anl.gov}

\author{Zhengying Liu}
\affiliation{%
 \institution{CNRS/Inria-LISN, U. Paris-Saclay}
   \country{France}
}
\email{zhengying.liu@inria.fr}

% \affiliation{%

% \affiliation{%

% \affiliation{%


\begin{abstract}
% \romain{word limit is 150, current is about 230}
Developing high-performing predictive models for large tabular data sets is a challenging task. Neural architecture search (NAS) is an AutoML approach that generates and evaluates multiple neural networks with different architectures concurrently to automatically discover an high performing model. A key issue in NAS, particularly for large data sets, is the large computation time required to evaluate each generated architecture. While data-parallel training has the potential to address this issue, a straightforward approach can result in significant loss of accuracy. To that end, we develop AgEBO-Tabular, which combines Aging Evolution (AE) to search over neural architectures and asynchronous Bayesian optimization (BO) to search over hyperparameters to adapt data-parallel training. We evaluate the efficacy of our approach on two large predictive modeling tabular data sets from the Exascale Computing Project-CANcer Distributed Learning Environment (ECP-CANDLE). 

\end{abstract}


\begin{comment}
\begin{CCSXML}
<ccs2012>
   <concept>
       <concept_id>10010147.10010257</concept_id>
       <concept_desc>Computing methodologies~Machine learning</concept_desc>
       <concept_significance>500</concept_significance>
       </concept>
   <concept>
       <concept_id>10010147.10010169.10010170</concept_id>
       <concept_desc>Computing methodologies~Parallel algorithms</concept_desc>
       <concept_significance>500</concept_significance>
       </concept>
   <concept>
       <concept_id>10010147.10010178.10010205</concept_id>
       <concept_desc>Computing methodologies~Search methodologies</concept_desc>
       <concept_significance>500</concept_significance>
       </concept>
 </ccs2012>
\end{CCSXML}
\end{comment}




\maketitle

\section{Introduction}

Tabular data sets are often diverse. They are obtained from multiple sources and modes, where combining certain inputs using \emph{problem-specific} domain knowledge typically leads to better and physically meaningful features and consequently robust models~\cite{erickson_autogluon-tabular_2020,zimmer_auto-pytorch_2020}. Many high-performing predictive models for tabular data are based on classical supervised machine learning (ML) methods such as bagging, boosting, and kernel-based methods. Specifically, ensemble methods that combine models obtained from different supervised ML methods have emerged as state-of-the-art for a wide range of predictive modeling tasks with tabular data. However, the design and development of such ensemble models is a highly iterative, manually intensive, and time-consuming task. Typically an ML pipeline for tabular data is composed of several components: data processing, dimension reduction, data balancing, feature selection, hyperparameter tuning, model selection, and ensemble strategy (such as stacking, bagging, and weighted combination). Given the design choices for each component, the complexity of designing an effective ML pipeline for tabular data is often beyond nonexperts.

Deep neural networks (DNNs) have achieved significant success in overcoming the issues of manual feature engineering and the complexities of developing a classical supervised ML pipeline. Nevertheless, designing DNNs for tabular data has received relatively less attention compared to image and text data. From the methodological perspective, there are two main reasons.

First, given the diversity of tabular data, designing DNNs with shared patterns such as convolutional and recurrent units is not meaningful unless further assumptions about the data are made. Second, fully connected DNNs, which are typically used for tabular data, can potentially lead to unsatisfactory performance because they can have large numbers of parameters, overfitting issues, and a difficult optimization landscape with low-performing local optima \cite{fernandez2014we}. 

Automated machine learning (AutoML) is a promising approach to address the methodological challenges in developing DNNs for tabular data. Neural architecture search (NAS), a class of AutoML, 
is an approach to automate the development of customized DNNs for a given data set. 
The NAS methods can be grouped into individual search methods and weight-sharing methods. The former generate a large number of architectures from a user-defined search space, train and validate each of them, and use the accuracy values to improve the generated architectures. 
The main advantage of these methods is parallelization: the generated architectures are independent, and they can be trained simultaneously. The disadvantage is that since each architecture is trained from scratch, architecture evaluation is expensive and becomes a bottleneck for effectiveness.   
To alleviate this issue, researchers proposed a different approach where the trained weights or computations are shared from an architecture to another during the search. This is enabled by defining a search space as an overparameterized network \cite{pham_efficient_2018} (also named hypernetwork), where the search samples subarchitectures and leverages the trained weights and computations from previously trained subarchitectures. This results in a significant reduction of evaluation time for several tasks. Nevertheless, the disadvantage of these methods is the instability due to the optimization gap between the supernetwork and its subarchitectures. In particular, optimizing the hypernetwork does not necessarily result in high-quality subarchitectures~\cite{chu_fair_2020}. 

We focus on individual NAS for large tabular data because of its ability to leverage multiple compute nodes to find high-performing neural networks. 
Specifically, we adopt aging evolution (AgE) \cite{real_regularized_2018}, a parallel NAS method that generates a population of neural architectures, trains them concurrently using multiple nodes, and improves the population by performing mutations on the existing architectures within a population. To reduce the training time of each architecture, we utilize the widely used distributed data-parallel training technique.

In this approach, the large training data is split into shards and distributed to multiple processing units. Multiple models with the same architecture are trained on different data shards, and the gradients from each model are averaged and used to update the weights of all models. Combining an individual NAS search method with distributed data-parallel training is a challenging task because the combination of the two methods requires nested parallelism. Moreover, the distributed data parallelism requires data-set-specific tuning of learning rate, and batch size in order to maintain accuracy and reduce training time. 

To that end, we make the following contributions:
\begin{itemize}
    %\item We show the negative impacts of data-parallelism in neural architecture.
    \item We develop AgEBO-Tabular, a joint neural architecture and hyperparameter search that combines aging evolution (AgE), a parallel NAS method \cite{real_regularized_2018} for searching the neural architecture space, and an asynchronous Bayesian optimization method for tuning the hyperparameters of data-parallel training. AgEBO-Tabular searches the architecture space and the hyperparameters of data-parallel training simultaneously. 
    \item We evaluate the efficacy of the proposed approach on two large tabular data sets from ECP-CANDLE benchmarks and show that AgEBO outperforms the accuracy of the AgE and discovers architectures that are faster to train. 
    \item We show that models produced by AgEBO outperform the manually designed models on the two ECP-CANDLE benchmark data sets.
\end{itemize}
The novelty of our work is fourfold: developing a new method for joint neural architecture and hyperparameter search, accelerating  NAS  with data-parallel training, using asynchronous Bayesian optimization for tuning the hyperparameters of data-parallel training,  and  advancing the state-of-the-art in the design of DNNs for large tabular data.

\section{Problem formulation}

Let $D_{train}$, $D_{valid}$, and $D_{test}$ be the training, validation, and test data, respectively. A neural architecture configuration $h_a$ is a vector from the neural architecture search space $H_a$, defined by a set of neural architecture decision variables. A hyperparameter configuration $h_m$ is a vector from hyperparameter search space $H_m$ defined by a set of hyperparameters. The joint neural architecture and hyperparameter search space $H$ is given by $H_a \times H_m$. The problem of joint neural architecture and hyperparameter search can be formulated as the following bilevel optimization problem:  
\begin{equation}
    \begin{aligned}
        &h_a^*, h_m^* = \argmax_{(h_a,h_m) \in H_a\times H_m} \mathcal{M}_{w^*}^{val}(h_a,h_m) \\
        &\st  w^* =  \argmin_{w} \mathcal{L}_{h_a,h_m}^{train}(w),
    \end{aligned}
    \label{eqn:pbopti3}
\end{equation}
where $\mathcal{M}_{w^*}^{val}(h_a,h_m)$ is the validation accuracy that needs to be maximized on $D_{valid}$ and $\mathcal{L}_{h_a,h_m}^{train}(w)$ is a loss function that needs to be minimised by optimizing the weights $w$ of the neural network configured with ($h_a$,$h_m$) using $D_{train}$. The test data $D_{test}$ is used only for the final evaluation.

The architecture search space differs from the hyperparameter search space with respect to the values that the decision variables take.  All the decision variables in the architecture search space belong to the categorical (nonordinal) type, where different values for a given variable do not have any particular order. On the other hand, the hyperparameter search space is characterized by mixed-integer variables. This comprises  integer, real, binary, and categorical types. Often, the number of categorical hyperparameters is relatively smaller than that of other types. Note that when all variables in the hyperparameter search space belong to a categorical type,  explicit partitioning in the search space is not required; consequently, a custom method such as our proposed AgEBO-Tabular for joint neural architecture and hyperparameter search becomes less relevant.

In our study, $H_a$ is defined by the decision variables to construct fully connected neural networks with skip connections for tabular data, and $H_m$ is defined by the hyperparameters of the data-parallel training (learning rate, batch size, optimizer, patience for learning rate reduction, patience for early stopping and loss function).

\section{AgEBO-Tabular}

The AgEBO-Tabular approach that we propose comprises three components: neural architecture search space  for tabular data, data-parallel training as evaluation strategy, and the AgEBO algorithm for joint neural architecture and hyperparameter search.

\subsection{Neural architecture search space for tabular data} 
\label{search space}

We model the search space of the neural architecture using a directed acyclic graph, which starts and ends with input and output nodes, respectively. They are fixed based on the input and output dimensions of the tabular data, respectively. It is possible to have multiple input nodes as well as output nodes. Between these, two sets of nodes are intermediate nodes, each of which can be a variable $\mathcal{N}$ or a skip-connection $\mathcal{SC}$ node. Each node represents a categorical decision variable that can take a list of nominal values (i.e., without order). 

Each variable node represents a dense layer with a list of different layer types; the choice is made by the NAS method.
The skip connections between the variable nodes are created by using skip-connection nodes. 
This type of node has two choices: zero for no skip connection and identity for the creation of skip connection.

\begin{figure}[!ht]
    \centering
    \begin{subfigure}{0.55\linewidth}
        \centering
        \includegraphics[width=1.\textwidth]{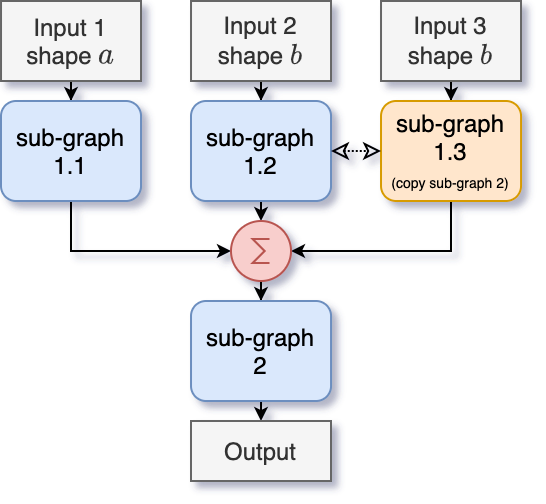}
        \caption{Global search space.}
        \label{fig:sub-graphs}
    \end{subfigure}
    \begin{subfigure}{0.65\linewidth}
        \centering
        \includegraphics[width=1.\textwidth]{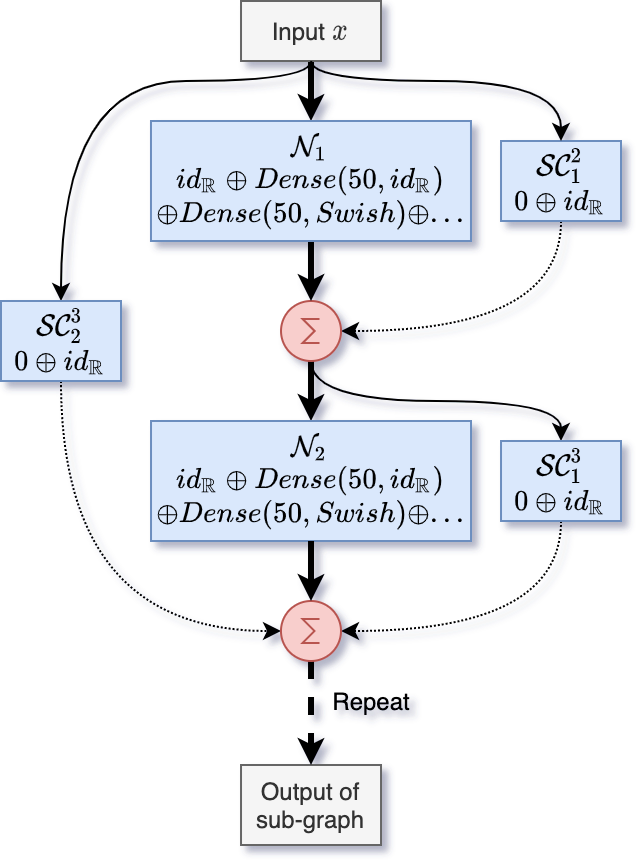}
        \caption{Detailed sub-graph.}
        \label{fig:specific-sub-graph}
    \end{subfigure}
    \caption{Neural architecture search space. The global search space is shown in~\ref{fig:sub-graphs}. The detailed structure of a sub-graph is shown in~\ref{fig:specific-sub-graph}. The nodes $\mathcal{N}_1$ and $\mathcal{N}_2$ represent  dense layers $Dense(x,y)$, where $x$ is the number of neurons and $y$ is the activation function. The nodes $\mathcal{SC}^2_1,\mathcal{SC}^3_1,\mathcal{SC}^3_2$ represent the possible skip-connection nodes, when $id_\mathbb{R}$ is chosen for each of them. The nodes shown in red with $\Sigma$ are used to represent the merging operators (sum or concatenation).}
    \label{fig:search_space}
\end{figure}

If multiple inputs are defined in the data, then subgraphs will be created for each of these inputs such as shown in Figure~\ref{fig:sub-graphs}. However, inputs with equal shape will benefit of parameter sharing and therefore be processed by the same subgraph.
 
Given a pair of consecutive variable nodes $\mathcal{N}_{k}$, $\mathcal{N}_{k+1}$, three skip-connection nodes $\mathcal{SC}^{k+1}_{k-3}, \mathcal{SC}^{k+1}_{k-2}, \mathcal{SC}^{k+1}_{k-1}$  are created. The choice of identity for these skip-connection nodes  respectively allows for connection to the three previous nonconsecutive variable nodes $\mathcal{N}_{k-3}, \mathcal{N}_{k-2}, \mathcal{N}_{k-1}$. 

For example, if an identity is chosen for $\mathcal{SC}^{k+1}_{k-1}$, a skip connection is made between $\mathcal{N}_{k-1}$ and $\mathcal{N}_{k+1}$ by passing the tensor output from $\mathcal{N}_{k-1}$ through a merging operator (e.g., concatenation, sum after projection or padding). 

In the case of sum, a linear layer is used to project the tensor from $\mathcal{N}_{k-1}$ to a correct shape. This is required for the creation of skip connections between $\mathcal{N}_{k-1}$ and $\mathcal{N}_{k+1}$ when their number of neuron units is different. The sum operator adds the projected input tensor from $\mathcal{N}_{k-1}$ and the tensor from $\mathcal{N}_{k}$, passes the summed tensor through the $ReLu$ activation function, and sends the resulting tensor as input to  $\mathcal{N}_{k+1}$. When $\mathcal{SC}^{k+1}_{k-2}$ and $\mathcal{SC}^{k+1}_{k-3}$ take identity values, the tensors from $\mathcal{N}_{k-2}$ and $\mathcal{N}_{k-3}$  undergo the same linear projection, and the tensor is given to the sum operator. When there is no skip connection, $\mathcal{SC}^{k+1}_{k-3}, \mathcal{SC}^{k+1}_{k-2}, \mathcal{SC}^{k+1}_{k-1}$ are set to zero; $\mathcal{N}_{k}$ and $\mathcal{N}_{k+1}$ are fully connected without the linear layer and the sum operator.
In the case of concatenation, the tensors are passed when there is a connection and simply concatenated together.

The same process is repeated for each of the $m$ variable nodes. See Figure~\ref{fig:specific-sub-graph} for an example.

The dense layer type is defined by the number of units and the activation function. For the former and the latter we used values in (50, 2000) with a step of 25 and a \{Identity, Swish \cite{ramachandran2018searching}, ReLu, Tanh, Sigmoid\}. These resulted in 391 (78 units $\times$ 5 activation functions, and identity) dense layer types for each variable node. Although one can order 391 values using the number of units in the layer, we did not consider and leverage such an order from the generality perspective. 

For example, if we consider only one value for the unit and different activation functions, then we cannot order the values in the list and cannot leverage the ordering in the NAS. We set the maximum number of variable nodes to $5$ for each sub-graph. The first variable node will not have a skip connection node. The second and the third variable nodes have 1 and 2 skip-connection nodes, respectively. The fourth to tenth variable nodes have 3 skip-connection nodes each.  The output node has 3 skip connections as well. 
%Consequently, the total number of architectures in the search space  is $31^{10} \times 2^{27} \approx 1.1 \times 10^{23}$.

\subsection{Data-parallel training as evaluation strategy} 
\label{section-data-parallelism}

The evaluation of an architecture in the individual NAS method consists of training the network and computing the validation accuracy. To speed up the evaluation, we use distributed data-parallel training. Given a neural architecture $\mathcal{A}$, the training data set is split in $n$ mutually exclusive subsets called shards, which are given to $n$ parallel processing units. Each of the $n$ processing units trains a copy of the same neural architecture $\mathcal{A}$ on its own shard. The gradients from each copy of neural architecture are synchronized and are used to update the weights. Moreover, we use the widely used linear scaling rule \cite{DBLP:journals/corr/GoyalDGNWKTJH17} to adapt the  learning rate and batch size depending on the level of parallelism in the data-parallel training. This heuristic states that the learning rate $lr_n$ and batch size $bs_n$ with $n$ processes should be scaled linearly with respect to $n$:
\begin{equation}
    lr_n = n * lr_1; bs_n = n*bs_1,
\end{equation}
where $lr_1,  bs_1$, are respectively the learning rate and batch size used for training with a single process. We treat $n$, $lr_1$, and $bs_1$ as hyperparameters and tune them using Bayesian optimization.
By leveraging the linear scaling rule, we try to achieve linear scaling for training time; however, there is an upper linear scaling limit above which the accuracy will suffer (without advanced and sophisticated layer-wise learning rate and adaptive batch size). Therefore, by tuning $n$, $lr_1$, and $bs_1$, we try to find the upper linear scaling limit that gives a maximal reduction in training time without losing accuracy.

\subsection{AgEBO: Aging evolution with Bayesian optimization} 
\label{agebo}

To perform a joint neural architecture and hyperparameter search, we propose aging evolution with Bayesian optimization (AgEBO). Our method combines AgE, a parallel NAS method for searching over the architecture space, and asynchronous Bayesian optimization (BO), for tuning the hyperparameters data-parallel training.

\edit{
There is not a lot of literature dedicated to asynchronous BO~\cite{pmlr-v97-alvi19a} compared to batch-synchronous BO~\cite{chevalier2013fast}. However, the asynchronous approach is justified by our practical case where the evaluated function can have significantly different runtimes depending on the architecture and hyperparameter configuration (see tables \ref{tab:best-networks-attn} and \ref{tab:best-networks-combo}). A way to easily parallelize BO is to use the \textit{constant liar} heuristic such as described in~\cite{hiot_kriging_2010,shahriari2015taking}.
% Asynchronous Bayesian optimisation~\cite{pmlr-v97-alvi19a} is of main importance to efficiently use distributed resources.
}

Algorithm \ref{alg:AgEBO} shows the pseudo code of AgEBO. The method follows the manager-worker paradigm for parallelization. It starts with $W$ workers, each with a maximum of $n_{max}$ parallel processing units for data-parallel training. The initialization phase starts by allocating an empty queue for the population of size $P$ and BO optimizer object. It is followed by sampling $W$ architecture configurations and hyperparameter configurations, respectively, and concatenating them. The neural network models are built by using the resulting configurations and are sent for concurrent evaluation on $W$ workers by using the submit\_evaluation interface (lines 3--7). Each worker uses the learning rate, batch size, and the number of processes from the configuration that it received to run the data-parallel training. The iterative part of the algorithm consists of collecting the results (validation accuracy values) once the workers finish their evaluation (line 9) and using them for generating the next set of architecture and hyperparameter configurations for evaluation. The BO optimizer object takes the hyperparameter configurations and their corresponding validation accuracy values and generates a $|results|$ number of hyperparameter values (using optimizer.tell and optimizer.ask interfaces, respectively, lines 12--13). To generate $|results|$ number of architecture configurations, the following steps are performed repeatedly: random sampling  $S$ architecture configurations from the incumbent population, selecting the best, and applying a random mutation to generate a child model hyperparameter configuration (lines 16--18). The generated architecture and  hyperparameter configurations are concatenated and sent for evaluation. Note that in the beginning of the search, the population queue does not have $P$ number of finished evaluations (given that all evaluations do not necessarily finish in the same time). Therefore, the architecture configurations are generated at random while the population size is smaller than $P$ (line 20)
The mutation corresponds to choosing a different operation for one variable node in the search space. This is achieved by first randomly selecting a variable node and then choosing (again at random) a value for that node excluding the current value. Then, the child is added to the population by replacing the oldest member of the population.

\begin{algorithm2e}[!ht]
\small
\DontPrintSemicolon
\SetInd{0.25em}{0.5em}
\SetAlgoLined
\SetKwInOut{Input}{inputs}\SetKwInOut{Output}{output}
\SetKwFunction{RandomPoint}{random\_point}
\SetKwFunction{SubmitEval}{submit\_evaluation}
\SetKwFunction{GetFinishedEval}{get\_finished\_evaluations}
\SetKwFunction{Push}{push}
\SetKwFunction{EmptyList}{EmptyList}
\SetKwFunction{RandomSample}{random\_sample}
\SetKwFunction{SelectParent}{select\_parent}
\SetKwFunction{Mutate}{mutate}
\SetKwFunction{Tell}{tell}
\SetKwFunction{Ask}{ask}

\SetKwFor{For}{for}{do}{end}

\Input{P: population size, S: sample size, W: workers}
\Output{highest-accuracy model in \textit{history}}
    \tcc{Initialization}
    $population \leftarrow$ create\_queue($P$) \tcp{Alloc empty Q of size P}
    {\color{blue} $optimizer \leftarrow$ optimizer()}\\
    
    \For{$i\leftarrow 1$ \KwTo $W$}{
        $model.h_a \leftarrow$ \RandomPoint{$H_a$}\\
        {\color{blue}$model.h_m \leftarrow$ \RandomPoint{$H_m$}}\\
        \SubmitEval{model} \tcp{Nonblocking}
    }
    \tcc{Main loop}
    \While{not done}{
        \tcp{Query results }
        $results \leftarrow$ \GetFinishedEval()\\
        \If{$|results| > 0$}{
            $population.$\Push{results} \tcp{Aging population}
            {\color{cyan} \tcp{Generate hyperparameter configs}}
            {\color{blue} $optimizer.$\Tell{$results.h_m, results.valid\_accuracy$}\\
            $next \leftarrow$ $optimizer.$\Ask{$|$results$|$}}\\
            \tcp{Generate architecture configs}
            \For{$i\leftarrow 1$ \KwTo $|results|$}{
                \eIf{$|population| = P$}{
                    $sample \leftarrow$ \RandomSample{population,S}\\
                    $parent \leftarrow$ \SelectParent{sample}\\
                    $child.h_a \leftarrow$ \Mutate{$parent.h_a$}\\
                }
                {
                $child.h_a \leftarrow$ \RandomPoint{$H_a$}\\
                }
                {\color{blue}$child.h_m \leftarrow next[i].h_m$}\\
                \SubmitEval{$child$} \tcp{Nonblocking}
            }
        }
    }

 \caption{AgE (black) and AgEBO (black + blue)}
 \label{alg:AgEBO}
\end{algorithm2e}

The BO component of AgEBO optimizes the hyperparameters ($h_m$) by marginalizing the architecture decision variables ($h_a$). The BO method generates hyperparameter configurations as follows. It starts by  sampling a large number of unevaluated hyperparameter configurations. 
For each sampled configuration $h_m^i$, it uses a model $M$ to predict a point estimate (mean value) $\mu(h_m^i)$ and standard deviation $\sigma(h_m^i)$. The sampled hyperparameter configurations are ranked by using the upper-confidence bound (UCB) acquisition function, \edit{an optimistic policy~\cite{shahriari2015taking} which consider the best case scenario in case of uncertainty}:
\begin{equation}
    UCB(h_m^i) = \mu(h_m^i) + \kappa\sigma(h_m^i),
    \label{eqn:ucb}
\end{equation}
where $\kappa \geq 0$ is a parameter that controls the trade-off between exploration and exploitation. A value of $\kappa=0$ corresponds to pure exploitation, where the hyperparameter configuration with the lowest mean value is always selected. A large value of $\kappa$ corresponds to exploration, where hyperparameter configurations with large variance are selected. Evaluation of such configurations results in  improvement of the model $M$. 
A typical BO optimization method with UCB is sequential and generates only one hyperparameter configuration at a time. This is not useful in our setting given the scale required by the AgE method. Therefore, to generate multiple hyperparameter configurations at the same time, we adopt an asynchronous BO that leverages multipoint acquisition function based on a constant liar strategy\edit{~\cite{hiot_kriging_2010}}. This approach starts by selecting a hyperparameter that maximizes the UCB function. The model $M$ is retrained with the selected hyperparameter configuration and a dummy value (lie) \edit{corresponding to the min value of collected objectives}. The next hyperparameter configuration is obtained by maximizing the UCB function using the updated model. The process of selecting a configuration and retraining the model with a lie is repeated until the required number of configurations are sampled. The mean of all the validation accuracy values found up to that point is used as a lie. While  several sophisticated asynchronous BO methods exist, the adoption of the constant liar strategy is motivated by its computational simplicity and low overhead. Since the mutation operation in AgE method is  simple, the BO method needs to generate multiple configurations in short computation time. Failure to do so will adversely affect the overall node utilization. This approach is motivated by the fact that we have more variables to optimize in the neural architecture space than in the hyperparameter space. Therefore, we can take advantage of the efficient sampling of BO without exploding the number of dimensions, which slow down the frequent refit of the surrogate model.

\subsection{Implementation details}

We implemented AgEBO in DeepHyper\footnote{\href{https://github.com/deephyper/deephyper}{https://github.com/deephyper/deephyper}}~\cite{deephyper_software},  open-source scalable AutoML software designed for neural architecture and hyperparameter search. A high-level implementation overview of the AgEBO method is shown in Figure \ref{fig:deephyper-arch}. We integrated the Ray Python package~\cite{moritz2018ray} within DeepHyper to schedule the evaluation of architectures concurrently. Specifically, at the beginning of each run, a cluster of Ray workers is launched. One of the nodes is called head node because it centralises the initial connections of all other nodes called workers. Each worker is launched with a set of available resources of 8 CPUs and 8 GPUs because Ray will enforce the isolation of available resources when tasks are submitted. Therefore if we define tasks which use 1 CPU and 1 GPU per task, we can place 8 tasks on this worker. Once the cluster is launched, we run our  Algorithm~\ref{alg:AgEBO} from the head node and call this process driver which connects to the Ray cluster. The function which evaluates architectures is exported as a Ray remote function with a maximum number of calls set to 1 to enforce a fresh restart of workers at each call and free properly the GPU memory reserved by Tensorflow. The evaluation function is also defined with a number of resources $R=(n_{cpu}, n_{gpu})$ needed for its execution. This number can vary in our study. For example, to train a neural network on 4 GPUs, we will set $R=(4,4)$ to the evaluation function. Therefore, the \texttt{submit\_evaluation} interface of AgEBO asks the Ray cluster to launch a task which is responsible for running the architecture training on $R$ resources, collecting the validation accuracy values, and returning the results through a \texttt{get\_finished\_evaluations} interface. All GPUs present on the head node are still available for computation.

\begin{figure}[!ht]
    \centering
    \includegraphics[width=0.8\linewidth]{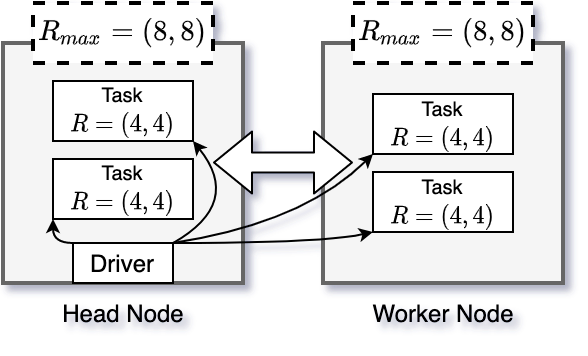}
    \caption{Overview of AgEBO implementation. The AgEBO search runs on a single process and uses the Ray workflow system to run the architecture evaluation on $W$ workers using the \texttt{ray.remote/get} interface.}
    \label{fig:deephyper-arch}
\end{figure}

\section{Experiments}

We used two large tabular data sets from the ECP-CANDLE pro\-ject~\cite{wozniak2018candle}: Combo and Attn. The selection was motivated by the unique nature of pharmacogenomic data. Both benchmarks consider the drug response problem, i.e., predicting the activity of a drug treatment against a cancer cell line, a critical step toward precision oncology. To accomplish this task, a host of diverse data types are used as input features. They include molecular assays such as protein, microRNA and gene expression profiles as well as drug descriptors and fingerprints. These tabular data modalities do not readily lend themselves to conventional deep learning architectures such as convolution. Yet, their intrinsic biological and chemical structures make them good candidates for mining inductive prior in model search. 
While Combo benchmark  models combinational drug response in a regression problem, and Attn is a single drug response classifier. Together, they include all five of the aforementioned tabular feature types. %Among all the data sets available in ECP-Candle Benchmark, we selected the following two large data sets:
\begin{enumerate}
    \item Combo~\cite{xia2018predicting} benchmark contains 220,890 data points in the training set and 55,222 data points in the testing set, 3 inputs with 942, 3,839 and 3,839 features respectively and 1 regression output. Compressed it takes about 4.2 GB in total. The task is to predict the growth percentage given a cell line molecular features and the descriptors of two drugs. We use 20\% of the training as a validation set because none is provided.
    \item Attn~\cite{clyde2020systematic} benchmark contains 271,915 data points in the training set, 33,989 in the validation and testing sets, 1 input with 6,212 features and 2 classes. Compressed it takes about 7.9 GB in total. The task is to classify the drug response in two classes. The distribution of classes in the data sets is unbalanced therefore we use the following class weights to mitigate this effect during learning: \{0: 0.52, 1: 13.87\}.
\end{enumerate}

For the Combo dataset, the baseline model is composed of two submodels followed by a final processing step. The first sub-model takes the cell expression as input. It is then composed of three dense layers each with 1,000 units and ReLU activation. The second sub-model has the same architecture, but it is used to process the two drug descriptors (i.e., weight sharing). Then the output of these submodels is concatenated and input in three dense layers with 1,000 units and ReLU activation to finally arrive to the output layer. The other hyperparameters are: mean squared error (MSE) for the loss, a batch size of 32, a learning rate of 0.01 and the optimizer is Adam.

For the Attn dataset, the baseline model is an attention-based neural network composed of 8 hidden dense layers of size [1000, 1000, 1000, 500, 250, 125, 60, 30] with ReLU activation function except for the third layer which has softmax activation due to the self-attention mechanism. Each of the dense layers is followed by batch normalization and dropouts of rate 0.2 are also used from the fourth layer. The other hyperparameters are:  categorical crossentropy for the loss, a batch size of 32, a learning rate of $10^{-5}$, the optimizer is SGD. A callback to reduce the learning rate on plateau is used with a patience of 40 and a factor of 0.2 while monitoring the validation AUROC of the model. A second callback for early stopping is also monitoring the validation AUROC and has a patience of 200.

To avoid overfitting, we did not use the test data set during the search. For every network generated during the search, the training data was used to train the model and the validation data was used to evaluate the accuracy. At the end of the search, we selected the best network based on the objective found at the last epoch of training, retrained it on the original training data, and evaluated its accuracy on the test data.

%and similar number of GPUs as during the search

Experiments were run on the ThetaGPU cluster at the Argonne Leadership Computing Facility (ALCF). ThetaGPU comprises 24 NVIDIA DGX A100 nodes, each equipped with eight NVIDIA A100 Tensor Core GPUs, two AMD Rome CPUs of 64 cores, 320 GB of GPU memory and 1 TB of DDR4 memory. 
\edit{Since there are already 8 GPUs per node, we did not consider the data-parallelism that spans across multiple nodes.
Instead, the data-parallel training within AgEBO is limited to a single node; however, it uses multiple GPUs within the single node to accelerate training. To optimise the training time and reduce overheads we cache the data in the DDR4 memory of each node. Whenever there are multiple evaluation tasks per node, they all access the data from the cache (no data set replication within each node)}.
We use the \texttt{cache} and \texttt{prefetch} methods from \texttt{tensorflow.dataset}, the \texttt{prefect} is set with \texttt{AUTOTUNE}. CUDA and CUDNN 11.0 are used jointly with NCCL 2.7.8. Finally, we activate the XLA compilation with \texttt{TF\_XLA\_FLAGS=--tf\_xla\_enable\_xla\_devices}.

By default, the NAS experiments were run for a wall time of 3 hours on 8 nodes of ThetaGPU. One process on the head node was reserved for the search, and all 64 GPUs were used as workers to train and validate the models generated by the search methods.

AgE was used as the baseline. The optimizer was set to Adam~\cite{kingma_adam_2017}, and each model was evaluated with a maximum of 100 epochs of training. The linear scaling rule~\cite{goyal_accurate_2018} was employed to scale the batch size and learning rate with respect to the number of parallel GPUs used for one evaluation. A callback was used to automatically reduce the learning rate on a plateau (\texttt{ReduceLROnPlateau}) with a patience of 5 epochs. An other callback was used to automatically stop the training (\texttt{EarlyStopping}) with a patience of 10 epochs. The objective is to maximise the validation $R^2$ for Combo and $AUC$ Precision-Recall for Attn. For the search, the population ($P$) and sample sizes ($S$) were set to $100$ and $10$, respectively. The batch size and learning rate were set to 32 and 0.001 (default values), respectively. AgEBO variants adopt the same training strategy as AgE uses. The difference between AgEBO variants and AgE is that the values of the batch size, learning rate, optimizer, loss, and the two patience can be tuned concurrently along with the architecture search. For AgEBO the range for hyperparameters was set as follows: batch size ($bs_1$) $\in$ $[|16,2048|]$; learning rate ($lr_1$) $\in$ (0.0001, 0.01), batch size and learning rate are sampled in a log-uniform scale within BO; optimizer $\in$  ["sgd", "rmsprop", "adagrad", "adam", "adadelta", "adamax", "nadam"], the patience of \texttt{ReduceLROnPlateau} $\in$ (3,30), the patience of \texttt{EarlyStopping} $\in$ (3,30). The loss $\in$ ["mae", "mse", "huber\_loss", "log\_cosh", "mape", "msle"] was used as a range for Combo because it was a regression problem.
Nevertheless, for the Attn, the loss was fixed to categorical cross-entropy. This is because we enforced exclusivity in the class label prediction using the softmax activation in the output layer with one-hot encoded target.

\subsection{Impact of hyperparameter tuning}

Here, we show that hyperparameter tuning with BO can significantly improve the accuracy of the neural architecture search. Moreover, we show that the accuracy of the architectures discovered by the AgE method with naive data-parallel training deteriorates significantly. Nevertheless, hyperparameter tuning in AgEBO circumvents this issue.

%nd varied the number of GPUs per architecture evaluation to analyze the time to solution and accuracy.

First, we evaluated AgE and AgEBO without data-parallel training. Next, we used data-parallel training, where we varied the number of GPUs per architecture evaluation for data-parallel training and analyzed time to solution and accuracy. We conducted all these experiments on 8 ThetaGPU nodes (default setting). We use \{AgE, AGEBO\}-$x$-$y$ to denote a variant, where $x$ and $y$ are the number of GPUs per architecture evaluation and the number of nodes, respectively. For example, AgE-1-8 refers to AgE ran with 1 GPU per architecture evaluation with 8 nodes. We used the default learning rate and batch size for AgE-1-8. Note that as we vary the number of GPUs per evaluation from 1, 2, 4, and 8, the number of simultaneous architecture evaluations become 64, 32, 16, and 8, respectively. In AgE-2-8, AgE-4-8, AgE-8-8, batch size and learning rate for different numbers of GPUs were scaled using the linear scaling rule. In AgEBO variants, all the hyperparameters are tuned using BO.

\begin{table}[!ht]
    \begin{subtable}[h]{1.\linewidth}
        \centering
        \resizebox{\linewidth}{!}{%
        \begin{tabular}{c|c|c|c|c}
        & \textbf{AgE-1-8} & \textbf{AgE-2-8} & \textbf{AgE-4-8} & \textbf{AgE-8-8} \\ \hline
         \textbf{\begin{tabular}[c]{@{}c@{}}Number of\\ architectures\end{tabular}} & 931            & 481           & 385           & 196 \\ \hline
        \textbf{\begin{tabular}[c]{@{}c@{}}\edit{Architectures}\\ training time (min.)\\ \edit{$\text{mean} \pm \text{std}$} \end{tabular}} & $11.51 \pm 6.35$ & $10.91 \pm 6.80$ & $6.79 \pm 4.79$ & $5.96 \pm 3.97$ \\ \hline
        \textbf{\edit{Best} Validation $R^2$} & 0.923 & 0.915 & 0.877 & 0.807
        \end{tabular}}
       \caption{}
        \label{tab:number-evaluated-architectures-age}
    \end{subtable}
    \hfill
    \begin{subtable}[h]{1.\linewidth}
        \centering
        \resizebox{\linewidth}{!}{%
        \begin{tabular}{c|c|c|c|c}
        & \textbf{AgEBO-1-8} & \textbf{AgEBO-2-8} & \textbf{AgEBO-4-8} & \textbf{AgEBO-8-8} \\ \hline
         \textbf{\begin{tabular}[c]{@{}c@{}}Number of\\ architectures\end{tabular}} & 1504            & 759           & 316           & 177 \\ \hline
         \textbf{\begin{tabular}[c]{@{}c@{}}\edit{Architectures}\\ training time (min.)\\ \edit{$\text{mean} \pm \text{std}$} \end{tabular}} & $7.05 \pm 4.77$ & $7.19 \pm 4.39$ & $8.54 \pm 5.02$ & $7.47 \pm 3.34$ \\ \hline
        \textbf{\edit{Best} Validation $R^2$} & 0.938 & 0.936 & 0.931 & 0.930
        \end{tabular}}
        \caption{}
        \label{tab:number-evaluated-architectures-agebo}
     \end{subtable}
     \caption{Results for data-parallel training in AgE (a) and AgEBO (b). 
        \edit{The mean and standard deviation (std) of architecture training time is computed from the training time of all the architectures found during the search. \vspace{-0.5cm}}}
     \label{tab:number-evaluated-architectures-age-agebo}
\end{table}

\begin{figure}[!h]
    \centering
    \begin{subfigure}{\linewidth}
        \centering
        \includegraphics[width=.8\textwidth]{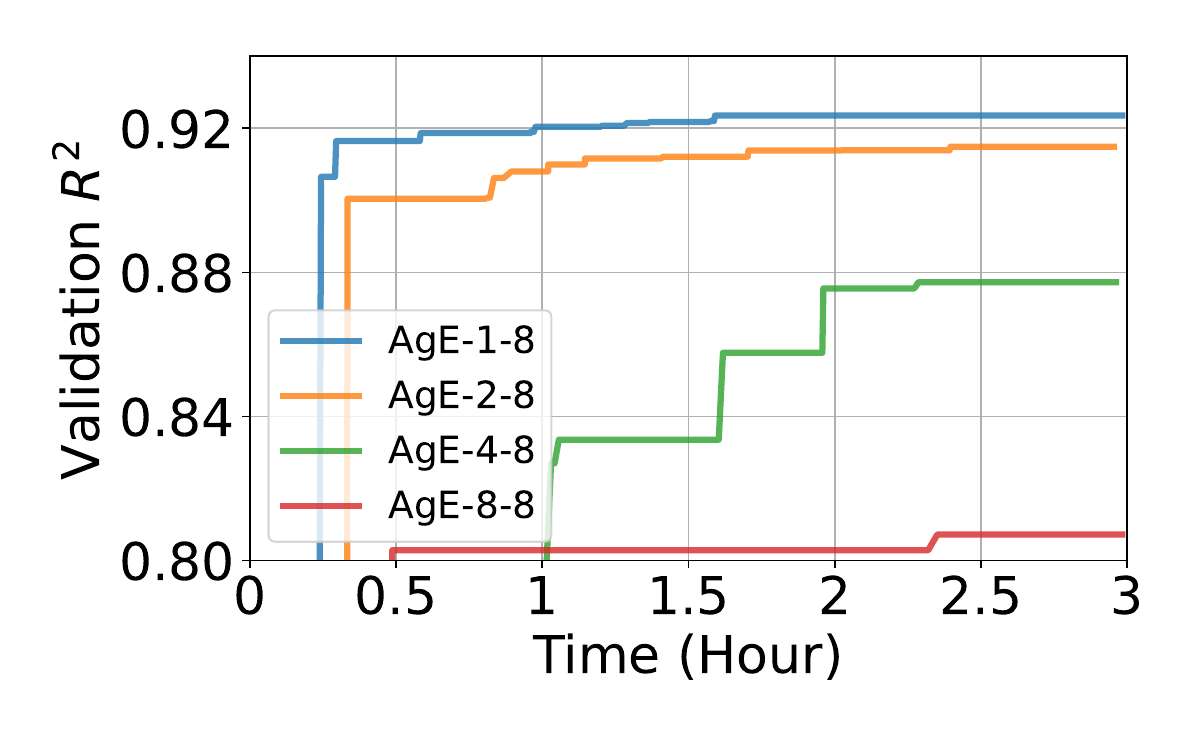}
        \caption{}
        \label{fig:best-objective-since-start-age}
    \end{subfigure}
    \begin{subfigure}{\linewidth}
        \centering
        \includegraphics[width=.8\textwidth]{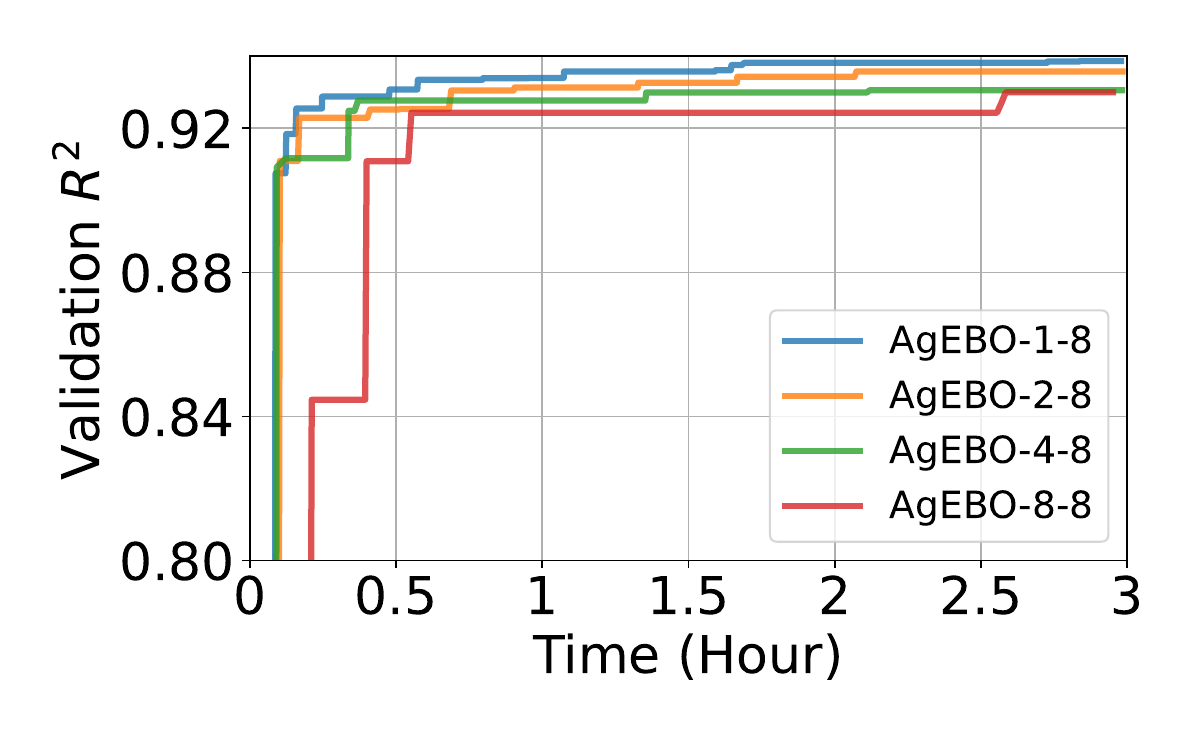}
        \caption{}
        \label{fig:best-objective-since-start-agebo}
    \end{subfigure}

    \caption{Search trajectory of AgE (\ref{fig:best-objective-since-start-age}) and AgEBO (\ref{fig:best-objective-since-start-agebo}) with different numbers of GPUs for data-parallel training on Combo data set. Each line denotes the best validation accuracy obtained over time. \vspace{-0.5cm}}
\end{figure}

The results of AgE and AgEBO variants on the Combo data set are shown in Figures~\ref{fig:best-objective-since-start-age} and \ref{fig:best-objective-since-start-agebo}. The plots show the validation $R^2$ of the best architecture found by the search over time.
 Tables~\ref{tab:number-evaluated-architectures-age} and \ref{tab:number-evaluated-architectures-agebo} show the number of architectures evaluated, training time, and validation $R^2$ of the best architecture found.

% The results of AgE on the Combo data set are shown in Figure~\ref{fig:best-objective-since-start-age}. The plot shows the validation $R^2$ of the best architecture found by the search over time.  Tables~\ref{tab:number-evaluated-architectures-age} and \ref{tab:number-evaluated-architectures-agebo} show the number of architectures evaluated, training time, and validation $R^2$ of the best architecture found. 

The comparison of AgE-1-8 and AgEBO-1-8 shows that the hyperparameter tuning with BO significantly improves the search to find high-performing architecture in a short time. AgE-1-8 reaches a validation $R^2$ of 0.923 around 100 minutes  whereas AgEBO-1-8 reaches that accuracy within 30 minutes and a validation $R^2$ of 0.938 at the end.

%we can observe that increasing the number of GPUs from 1 to 8 per architecture evaluation decreases the validation $R^2$ from 0.923 to 0.807. 
 
From the results of AgE-2-8, AgE-4-8, AgE-8-8, we can see that the naive data-parallel training significantly affects the AgE's ability to find architectures with high accuracy. The training time distribution computed from all architectures found by the search shows a significant reduction in training time from 2 to 4 GPUs, but the speedup from the use of 8 GPUs is negligible. We did not observe a significant reduction in training time from 1 GPU to 2 GPUs. This is because AgE-1-8 was run without Tensorflow distributed library but AgE-2-8 (AgE-4-8 and AgE-8-8) incurs the overhead of using it. We can also observe that the number of architectures explored by AgE reduces with an increase in the number of GPUs per architecture evaluation. All these observations clearly establish that the naive data-parallel training degrades the performance of AgE.

The results of AgEBO-2-8, AgEBO-4-8, AgEBO-8-8 that have autotuned data-parallel training show that they reach similar accuracy for different number of GPUs per architecture evaluation. AgEBO-2-8, AgEBO-4-8, AgEBO-8-8 reach the accuracy of AgE-1-8 within 40 minutes of search. There is a slight reduction in accuracy (to the third decimal place) as we increase the number of GPUs, but the reduction is not as drastic as seen with AgE variants. For the same number of GPUs per architecture evaluation, AgEBO achieves better validation $R^2$ than that of AgE. These results show that the joint optimization of hyperparameters and neural architectures is able to circumvent the issues posed by the naive data-parallel training. It is interesting to note that despite the significant reduction in the number of architectures evaluated, AgEBO counters the loss of accuracy by tuning the hyperparameters, which eventually leads to a similar training time distribution. AgE alone cannot achieve this with the fixed hyperparameter values during the search.

\subsection{Comparison with the manually-designed baseline}

Here, we show that the best models found by AgEBO outperform the manually designed neural network baselines for Combo and Attn with respect to accuracy and training time. 

%In this part of the study we demonstrate the consistency of our approach on 2 data sets without tuning the search space or the hyperparameters of AgEBO (e.g., population size, exploration of BO). 

For the comparison, we selected the best model obtained by AgEBO and retrained the model on a single ThetaGPU node with a wall-time of 1 hour and a maximum of 100 epochs. We also included the best model found by AgE-1-8 (AgE without data-parallel training) for comparison.

\begin{table}[!ht]
    \begin{subtable}[h]{\linewidth}
        \centering
        \begin{tabular}{c|c|c|c|c|c}
                          & \textbf{\begin{tabular}[c]{@{}c@{}}Number\\ of\\ Param.\end{tabular}} & \textbf{\begin{tabular}[c]{@{}c@{}}Training\\ Time\\ (min.)\end{tabular}} & \textbf{\begin{tabular}[c]{@{}c@{}}Test\\ MSE\end{tabular}} & \textbf{\begin{tabular}[c]{@{}c@{}}Test\\ MAE\end{tabular}} & \textbf{\begin{tabular}[c]{@{}c@{}}Test\\ $R^2$\end{tabular}} \\ \hline
        \textbf{Baseline} & 13,791 K & 35.42   & 0.0249  & 0.1051  & 0.902      \\ \hline
        \textbf{AgE-1-8} & 15,328 K  & 12.01  & 0.021 & 0.092  & 0.919   \\ \hline
        \textbf{AgEBO-1-8}    & 37,688 K  & 9.50  & 0.018  & 0.080  & 0.931   \\ \hline           
        \textbf{AgEBO-2-8}    & 25,518 K  & 4.88  & 0.019  & 0.082  & 0.928   \\ \hline           
        \textbf{AgEBO-4-8}    & 29,266 K  & 9.12  & 0.020  & 0.086  & 0.921   \\ \hline
        \textbf{AgEBO-8-8}    & 14,263 K  & 10.25  & 0.020  & 0.086  & 0.920  
        \end{tabular}
        \caption{}
        \label{tab:best-networks-combo}
    \end{subtable}
    \hfill
    \begin{subtable}[h]{\linewidth}
        \centering
        \begin{tabular}{c|c|c|c|c|c}
                          & \textbf{\begin{tabular}[c]{@{}c@{}}Number \\ of\\ Param\end{tabular}} & \textbf{\begin{tabular}[c]{@{}c@{}}Training\\ Time\\ (min.)\end{tabular}} & \textbf{\begin{tabular}[c]{@{}c@{}}Test\\ Loss\\\end{tabular}} & \textbf{\begin{tabular}[c]{@{}c@{}}Test\\ AUC\\ ROC\end{tabular}} & \textbf{\begin{tabular}[c]{@{}c@{}}Test\\ AUC\\ P-R\end{tabular}} \\ \hline
        \textbf{Baseline} & 8,893 K                                                      & 60.40                                                                     & 0.141                                                                              & 0.984                                                             & 0.977                                                             \\ \hline
        \textbf{AgE-1-8}    & 13,933 K                                                              & 8.77                                                                      & 0.225                                                                              & 0.973                                                             & 0.970                                                             \\ \hline
        \textbf{AgEBO-1-8}    & 33,866 K                                                              & 2.51                                                                      & 0.088                                                                              & 0.995                                                             & 0.995                                                            \\ \hline
        \textbf{AgEBO-2-8}    & 42,145 K                                                              & 1.49                                                                      & 0.116                                                                              & 0.992                                                             & 0.992                                                             \\ \hline
        \textbf{AgEBO-4-8}  & 46,561 K                                                              & 4.20                                                            & 0.078                                                                     & 0.996                                                    & 0.996                                                    \\ \hline
        \textbf{AgEBO-8-8}  & 40,153 K                                                              & 5.13                                                                      & 0.090                                                                              & 0.995                                                             & 0.994                                                            
        \end{tabular}
        \caption{}
        \label{tab:best-networks-attn}
     \end{subtable}
     \caption{Metrics for the best models obtained different strategies for Combo (a) and Attn (b). }
\end{table}

\begin{figure}[!h]
    \centering
    \begin{subfigure}{\linewidth}
        \centering
        \includegraphics[width=.8\textwidth]{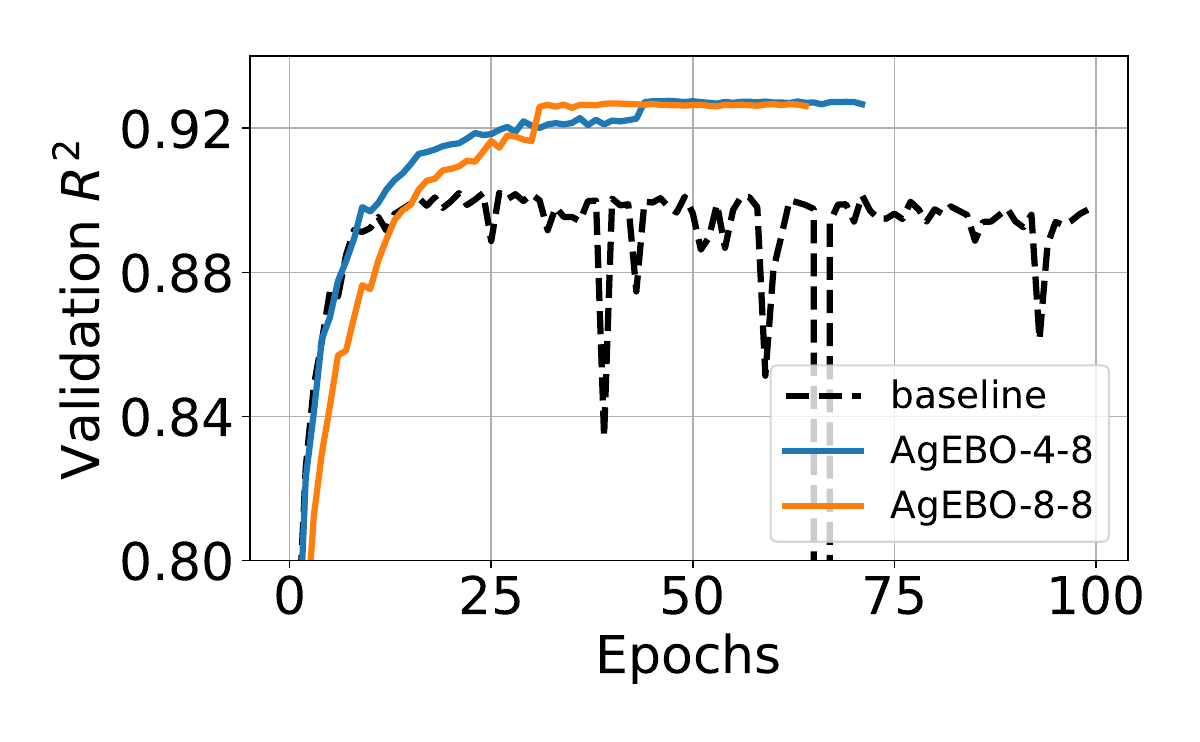}
        \caption{Combo}
        \label{fig:best-networks-combo}
    \end{subfigure}
    \begin{subfigure}{\linewidth}
        \centering
        \includegraphics[width=.8\textwidth]{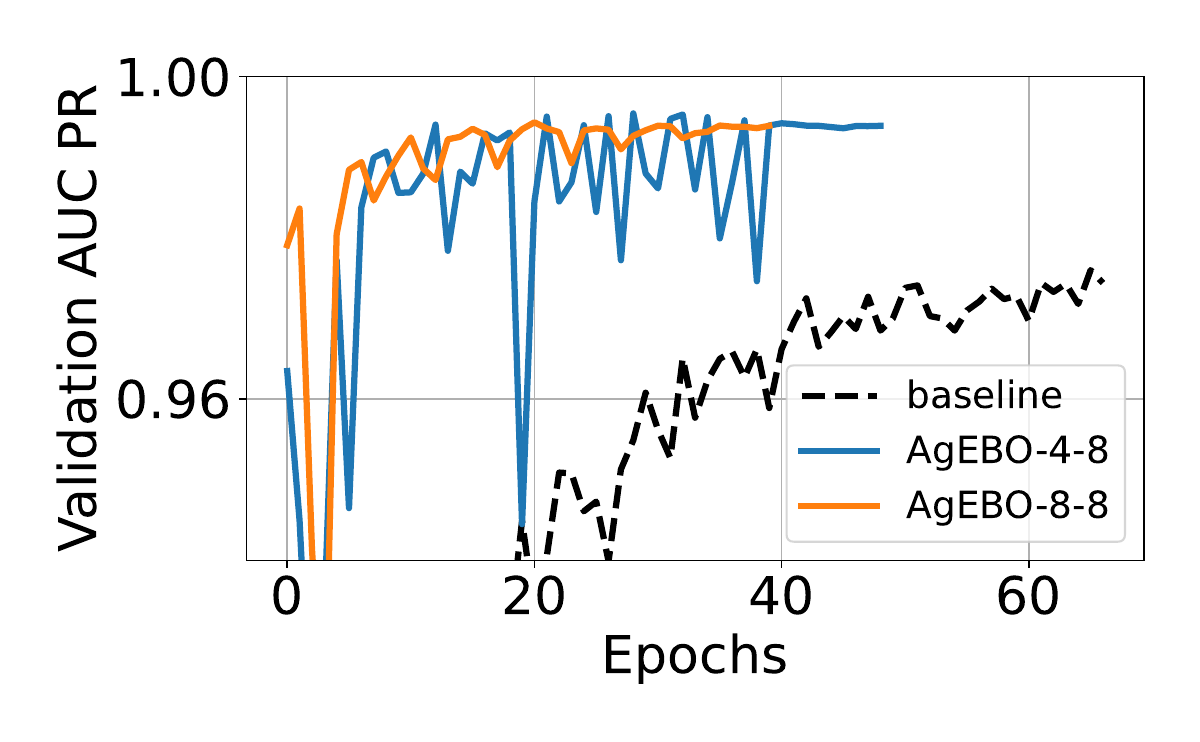}
        \caption{Attn}
        \label{fig:best-networks-attn}
    \end{subfigure}

    \caption{Training profiles of the best models found by AgEBO-4-8, AgEBO-8-8, and the baseline \edit{(dashed line)}. \vspace{-0.5cm}}
    \label{fig:best-networks}
\end{figure}

The results are shown in Tables~\ref{tab:best-networks-combo} and~\ref{tab:best-networks-attn}. From the results, we can observe that AgEBO variants outperform both Combo and Attn baselines with respect to all the  metrics. We observe that there is a slight decrease in the accuracy values by using the best found models from AgEBO-2-8, AgEBO-4-8, and AgEBO-8-8. For both data sets, AgEBO discovers network architectures with large number of trainable parameters to increase the accuracy but at the same time they are much faster to train. For Combo, the best network  found by AgEBO-2-8 is 7.25x faster than the baseline. For Attn, the best network found by \edit{AgEBO-2-8 is 40.53x} faster than the baseline. The evolution of the validation $R^2$ for a few best performing network is shown in Figure~\ref{fig:best-networks-combo}. For Combo, the baseline starts to stagnate and oscillate from 25 epochs. The best network found by AgEBO reaches a high accuracy within 25 epochs. For Attn, which is characterised by heavy class imbalance, the baseline improves slowly and and stops after 70 epochs because of the wall time. The best network found by AgEBO reaches a high accuracy within 20 epochs, oscillates from 20 to 40 trying to improve, and stops without improvement after 50 epochs.   

%The starts to stagnate at at a much lower plateau than NAS generated networks. Also, its training does not stop before reaching the maximum number of epochs even if the performance does not improve.

%The results are shown in Table~\ref{tab:best-networks-combo} and Figure~\ref{fig:best-networks-combo}. The baseline has an $R^2$ of 0.902 and about 14K parameters which is the lower number of parameters compared to all the models generated by either AgE or AgEBO. It is clear that all the AgEBO variants are able to improve over AgE. Also, the fastest training network takes about 5 minutes which is a 7 times faster than the baseline. This speed-up is interesting to retrain the model regularly with more data. The best $R^2$ reach is 0.931 which is also a significant improvement over the baseline. 

\edit{The key advantages of AgEBO in light of multiple GPUs per evaluation are high accuracy and faster training of the best model found during the search. This is evident from the results of Table 3 and 4. Given the same number of nodes, AgEBO-2-8 achieves models with training times that are 1.94x (Combo) and 1.68x (Attn) faster than AgEBO-1-8 (Table 3 and Table 4) without significant loss in accuracy.  AgEBO-2-8 starts with 2 GPUs per evaluation, which can result in faster models during the initial iterations of the search but their accuracy values will be low. As the search proceeds, due to the asynchronous nature of AgEBO, the models with high validation accuracy values that train faster are reinforced and have a higher chance of survival in the population. However, further increases in the number of GPUs per evaluation do not result in significant reduction in training time for these data sets. To offset the loss of accuracy in AgEBO-4-8 and AgEBO-8-8, AgEBO generates models that take longer to train (more epochs to reach similar accuracy). The comparison between AgE-1-8 and AgEBO-1-8 shows that even in the absence of data parallel training, optimizing the hyperparameters through BO increases the accuracy and decreases the training time. For Combo and  Attn data sets, the test $R^2$ values improve from 0.919 to 0.931 and from 0.970 to 0.995, respectively. The training times of the best models from AgEBO-1-8 are 1.2x and 3.49x faster than that of AgE-1-8.} It is worthwhile to mention that faster training and inference models are useful for a number of downstream cancer predictive modelling tasks, such as training a large number of models with the same architecture but with different random seeds for building a robust ensemble with uncertainty quantification and accelerating high-throughput in silico drug pair screening with faster inference.    

%The results are shown in Table ~\ref{tab:best-networks-attn} and Figure~\ref{fig:best-networks-attn}. The baseline has loss, AUC ROC and AUC Precision-Recall of respectively 0.141, 0.984 and 0.977. The best model found is trained in about 4 minutes a 14 times speed-up compared to baseline. Also, it reach a loss of 0.078, an AUC ROC of 0.996 and a AUC Precision-Recall of 0.996 which are all better than baseline.

\subsection{Scaling}

Here, we compare the scaling behaviour of AgE and AgEBO and show that hyperparameter tuning helps AgEBO to achieve better scaling.

All the experiments were run with a wall time of 3 hours and we use 2 GPUs per architecture evaluation. The number of nodes is progressively increased from 1, 2, 4, 8, and 16. Given that in these settings, we have 4 workers per node (8 GPUs for each node with 2 GPUs/evaluation), it results in a minimum of 4 workers (1 node) and a maximum of 64 workers (16 nodes) in parallel.

\begin{figure}[!ht]
    \centering
    \includegraphics[width=0.8\linewidth]{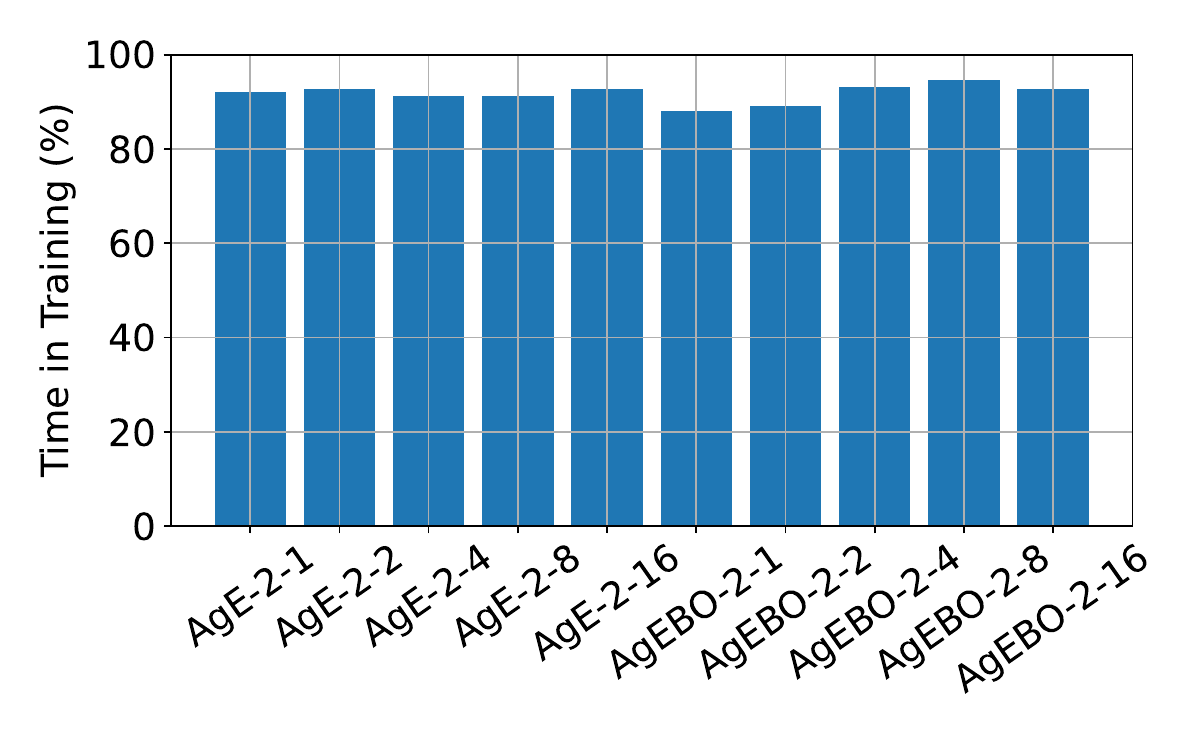}
    \caption{Normalized cumulative time spent in training neural networks for AgE and AgEBO variants. \vspace{-0.5cm}}
    \label{fig:strong-scaling-training-time}
\end{figure}

Figure~\ref{fig:strong-scaling-training-time} shows the percentage of cumulative time spent in training the generated neural networks normalized by the total available time of GPU computation (i.e., $3 \text{ hours} \times \text{number of  GPUs}$). We observe that minimal time is spent in starting the Ray cluster, loading the data or initialising the model before training. These results show that AgEBO does not have significant overhead to generate networks for evaluation given the node counts considered. Note that these measurements do not take into account the overhead incurred by Tensorflow distributed or gradient synchronisation. 

Table~\ref{tab:strong-scaling} summarizes the scaling results. For AgEBO, we can observe that increasing the number of nodes increases the number of evaluations and improves the best validation $R^2$ value. More importantly, the time to reach baseline accuracy significantly reduces by increasing the number of nodes. By scaling to 4 and above nodes, AgEBO reaches the accuracy of the baseline with 7 minutes. While the trends are similar for AgE, the number of evaluations, the best validation $R^2$ value, and the time to reach baseline are poorer than those of AgEBO. These results clearly show that BO significantly helps AgEBO in scaling.

%At 16 nodes, there is no significant reduction in the time. 
%that scaling helps to reduce the time required to reach a configuration at least as good as the baseline.  Moreover, the best validation $R^2$ also improves with the number of nodes. Initially of 0.768 (1 node) to 0.915 (8 nodes) with AgE it significantly improves from 0.915 (1 node) to 0.931 (8 nodes) with AgEBO. Now, for the training time looking at the standard deviation from the mean it is clear that networks explored during the search are having very different training durations.

\begin{figure}[!ht]
    \centering
    \includegraphics[width=0.8\linewidth]{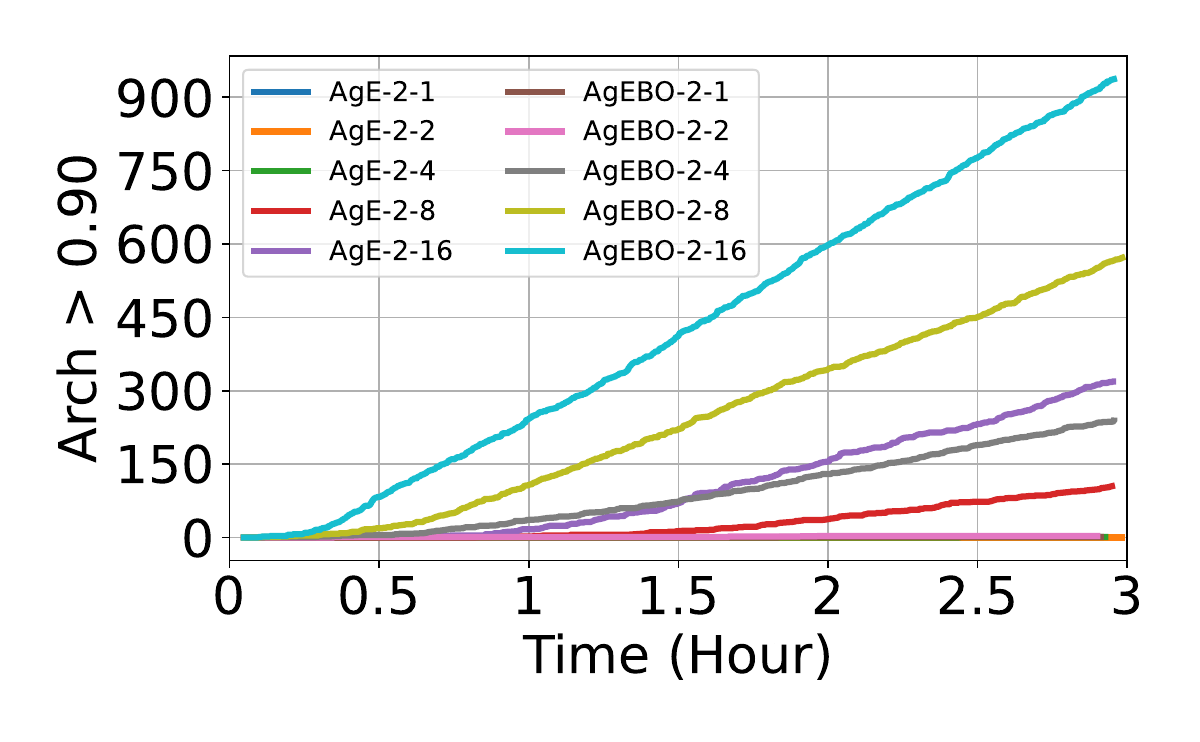}
    \caption{Number of unique architectures found by AgE and AgEBO variants that are better than the manually-designed baseline. \vspace{-0.4cm}}
    \label{fig:strong-scaling-unique-arch}
\end{figure}

The results in Table~\ref{tab:strong-scaling} \edit{do} not capture the usefulness of the scaling in its entirety. To that end, we analyzed the number of unique architectures obtained by AgE and AgEBO that are better than the baseline over time. This measures the strength of AgE and AgEBO to outperform the baseline and how it changes as we increase the number of nodes. The results are shown in Figure~\ref{fig:strong-scaling-unique-arch}. These results clearly show that AgEBO and AgE take advantage of large number of nodes to find architectures that are better than baseline. For AgEBO, we found a linear scaling for up to 8 nodes: the number of unique architectures that are better than baseline almost doubles by doubling the number of nodes. At 16 nodes, there is a drop in linear scaling.
Furthermore, we can observe that AgEBO completely dominates AgE. The best AgE variant, AgE-2-16 obtains 300 unique architectures that are better than baseline after three hours. The best AgEBO variant, AgEBO-2-16 reaches the same number in 75 minutes and in three hours it found more than 800 architectures that are better than the baseline. 
\vspace{-0.5cm}

%It is clear that from 8 nodes AgE and AgEBO produces a lot more good configurations. Also, for a fixed number of nodes the AgEBO case is better than AgE. For example, in the 8 nodes case AgEBO produces about twice more good configurations than AgE.

\begin{table}[!h]
% \resizebox{\columnwidth}{!}{%
\begin{tabular}{c|c|c|c}
                    & \textbf{\begin{tabular}[c]{@{}c@{}}Number\\ of\\ Evaluation\end{tabular}} & \textbf{\begin{tabular}[c]{@{}c@{}}Best\\ Validation\\ R2\end{tabular}} & \textbf{\begin{tabular}[c]{@{}c@{}}Time to\\ Baseline\\  (min.)\end{tabular}} \\ \hline
\textbf{AgE-2-1}    & 78 & 0.768 & -                                                                           \\ \hline
\textbf{AgE-2-2}    & 145 & 0.830  & -                                                                           \\ \hline
\textbf{AgE-2-4}    & 246                                                                                                                                     & 0.904                                                                   & 146.73                                                                        \\ \hline
\textbf{AgE-2-8}    & 481                                                                                                                                     & 0.915                                                                   & 49.75                                                                         \\ \hline
\textbf{AgE-2-16}   & 983                                                                                                                                  & 0.919                                                                      & 23.62                                                                            \\ \hline \hline
\textbf{AgEBO-2-1}  & 88                                                                                                                                      & 0.915                                                                   & 110.27                                                                        \\ \hline
\textbf{AgEBO-2-2}  & 88                                                                                                                                    & 0.911                                                                   & 21.06                                                                         \\ \hline
\textbf{AgEBO-2-4}  & 331                                                                                                                                     & 0.934                                                                   & 6.96                                                                          \\ \hline
\textbf{AgEBO-2-8}  & 759                                                                                                                                     & 0.936                                                                   & 6.15                                                                          \\ \hline
\textbf{AgEBO-2-16} & 1196                                                                                                                                               & 0.936                                                                      & 6.42                                                                           
\end{tabular}
% }
\caption{Scaling the number of nodes with 2 GPUs per architecture evaluation for AgE and AgEBO. A "-" denotes that the condition was not met. \vspace{-1cm}}
\label{tab:strong-scaling}
\end{table}

\subsection{Comparison with mixed BO and mixed AgE}
\label{mixed-age}

In AgEBO tabular, the BO component of AgEBO optimizes the hyperparameters by marginalizing the architecture decision variables. Here, we analyze if this is an effective strategy when compared to BO that optimizes hyperparameters along with architecture decision variables and AgE that optimizes architecture decision variables along with hyperparameters. We refer these two methods as mixed BO and mixed AgE, respectively. 

In mixed BO, we used the same asynchronous Bayesian optimization based on LCB acquisition function and with the same $\kappa=0.001$. In mixed AgE, we concatenate the list of hyperparameters and neural architecture discrete dimensions. At each iteration of AgE, one of these dimensions is picked at random and the mutation corresponds to a sampling from the prior distribution of this selected dimension. These experiments are run on the Combo data set with 8 nodes and 4 GPUs per evaluation.

\begin{figure}[!t]
    \centering
    \begin{subfigure}{\linewidth}
        \centering
        \includegraphics[width=.8\textwidth]{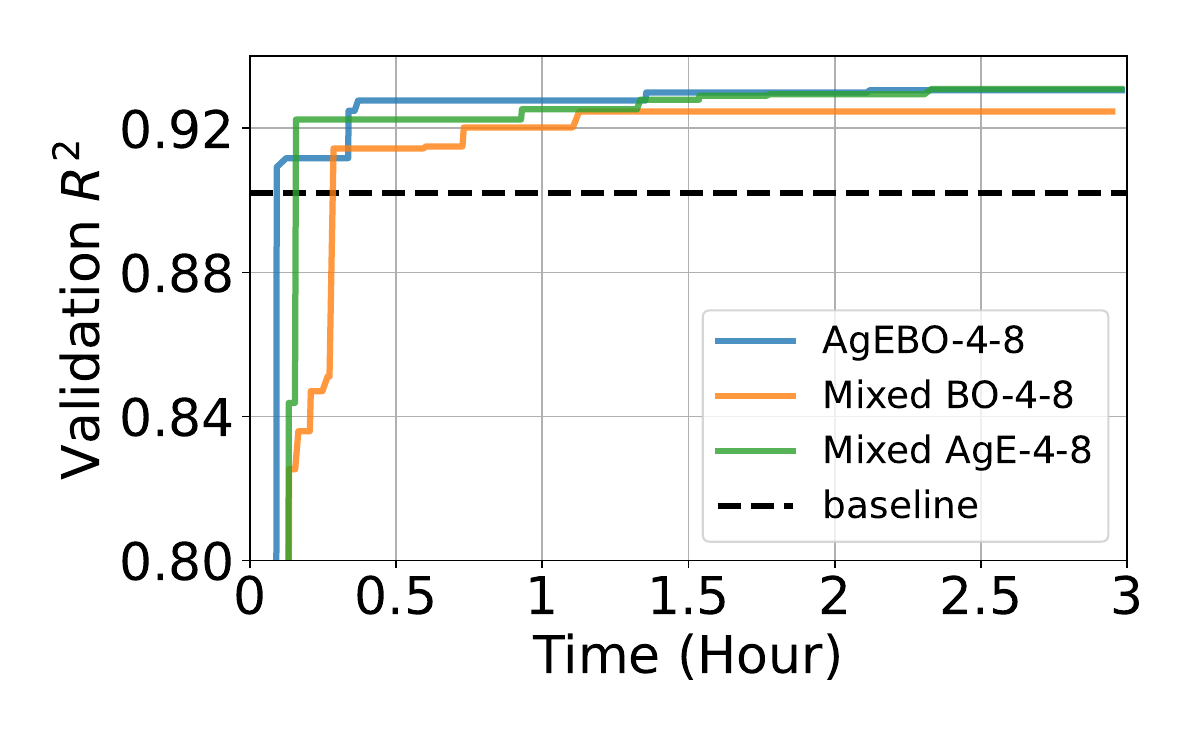}
        \caption{}
        \label{fig:agebo-bo-age-objective}
    \end{subfigure}
    \begin{subfigure}{\linewidth}
        \centering
        \includegraphics[width=.8\textwidth]{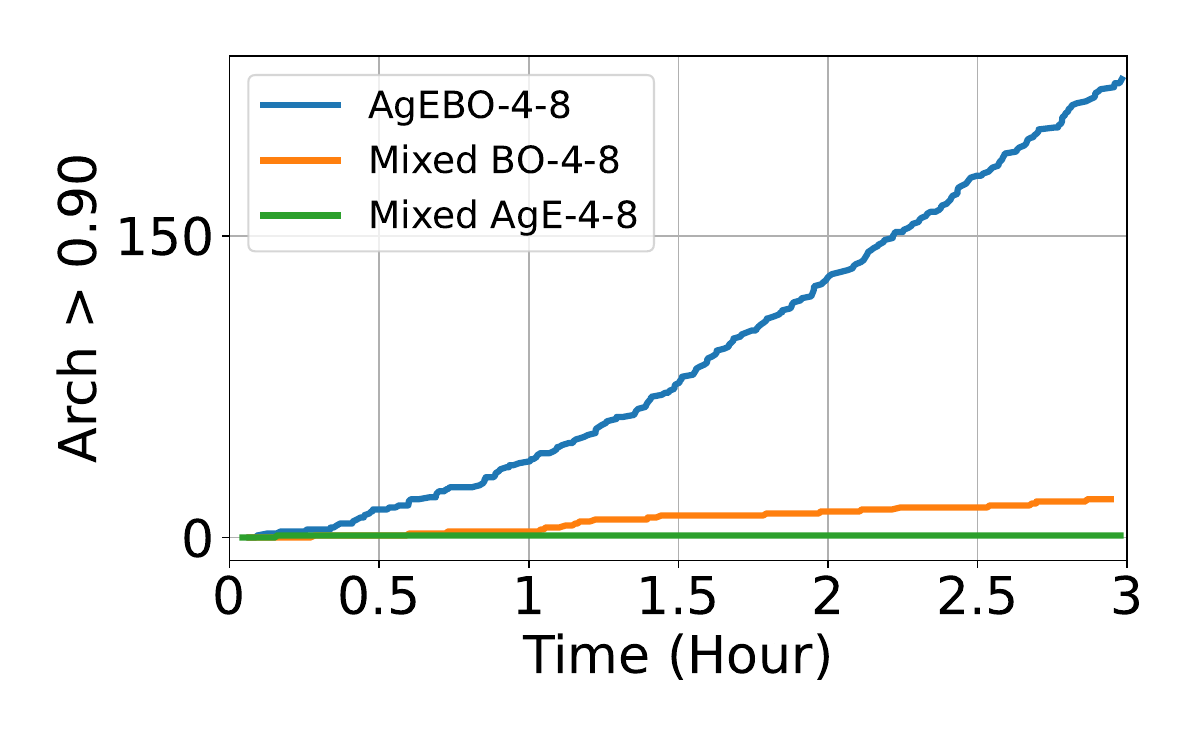}
        \caption{}
        \label{fig:agebo-bo-age-unique-architectures}
    \end{subfigure}
\vspace{-0.4cm}
    \caption{Comparison of AgEBO with mixed AgE and mixed BO on the Combo data set. (a) Search trajectory showing the best validation $R^2$ found by the search methods obtained over time (b) the number of unique architectures found by the search methods that are better than the baseline. \vspace{-0.8cm}}
    \label{fig:agebo-bo-age}
\end{figure}

The results are shown in Figure~\ref{fig:agebo-bo-age-objective}. We can observe that AgEBO-4-8 and mixed AgE-4-8 outperform mixed BO-4-8 with respect to validation $R^2$ values. While AgEBO and mixed AgE reach 0.931, BO stagnates at 0.925. Figure~\ref{fig:agebo-bo-age-unique-architectures} shows the number of unique  architectures that are better than the baseline found by the three methods over time. We can observe that AgEBO outperforms both mixed AgE and BO. In 15 minutes, AgEBO achieves the number that is better than both mixed AgE and BO and it obtains more than 200 unique architectures that are better than the baseline.

\edit{The effectiveness of marginalization in AgEBO can be attributed to the implicit regularization mechanism of the AgE method and a stronger exploitation of the BO method. The AgE's regularization mechanism favours architectures that retrain well across multiple generations. The only way  architecture can stay in the population for a long time is through inheritance---it should be passed down from parent to child through multiple generations. Whenever an architecture is inherited it will undergo retraining. If the retraining validation accuracy becomes low, the architecture will be removed from the population. Given that BO has a stronger exploitation due to small $\kappa$ value, the hyperparameter values generated will be biased towards the hyperparameter values of the previously obtained high-performing architectures in the population. If any of the high-performing architectures becomes low-performing after inheritance and retraining, it will be removed from the population. Consequently, only the architectures that result in the improvement of the validation accuracy after inheritance and retraining are allowed to evolve in the population during the search.}

\subsection{Exploration and exploitation in AgEBO}

Here, we study the effect of exploration and exploitation of BO within AgEBO by varying $\kappa$ values. We show that stronger exploitation is critical for the effectiveness of AgEBO.

The $\kappa$ value in Eq.~\ref{eqn:ucb} controls the trade-off between exploration and exploitation in BO. In addition to the default $\kappa$ value of 0.001, we ran AgEBO with \edit{six values: \{0, 0.001, 0.01, 0.1, 1.96, 19.6\}}. Note that $1.96$ is the typical $\kappa$ value in Scikit-Optimize, which provides a balance between exploration and exploitation. The value of 19.6 is selected to enforce large exploration. \edit{On the other hand, the value of 0 is to enforce pure exploitation where the variance in Eq.~\ref{eqn:ucb} is totally ignored. The values of \{0.01, 0.1\} enforce different degrees of exploitation.} We ran the experiments on the Combo data set.

\begin{figure}[!h]
    \centering
    \begin{subfigure}{\linewidth}
        \centering
        \includegraphics[width=.8\textwidth]{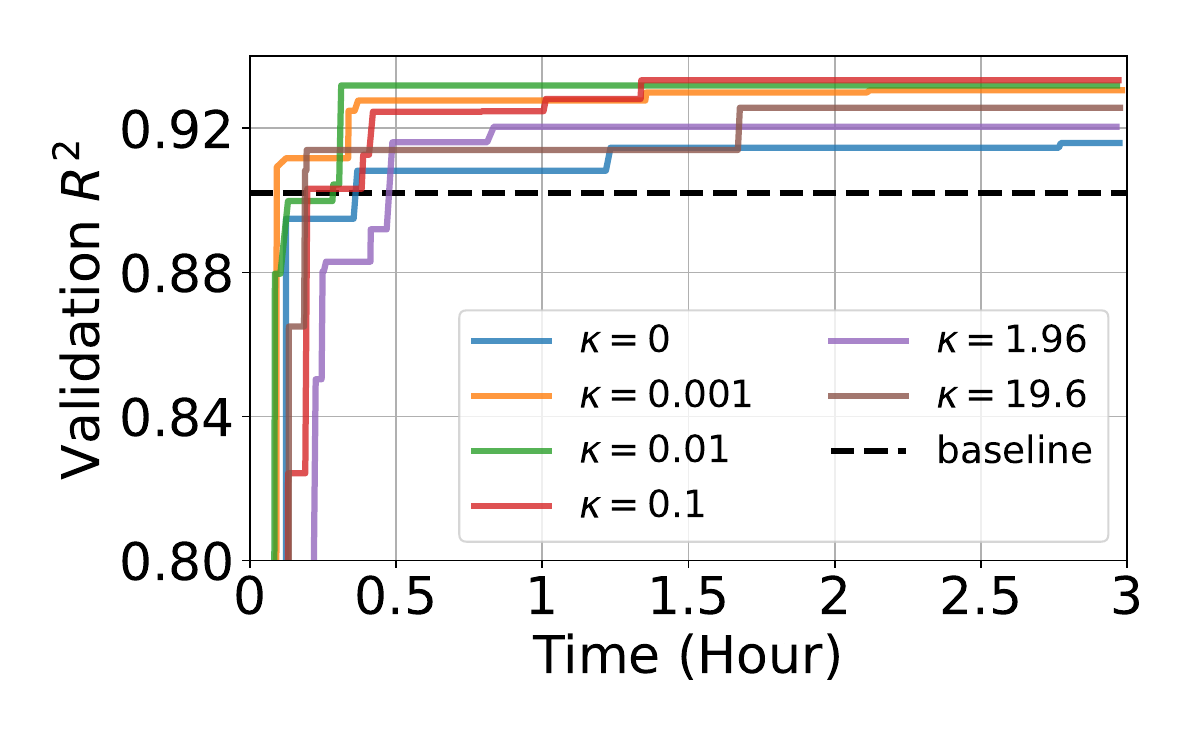}
        \caption{}
        \label{fig:agebo-exploration-objective}
    \end{subfigure}
    \begin{subfigure}{\linewidth}
        \centering
        \includegraphics[width=.8\textwidth]{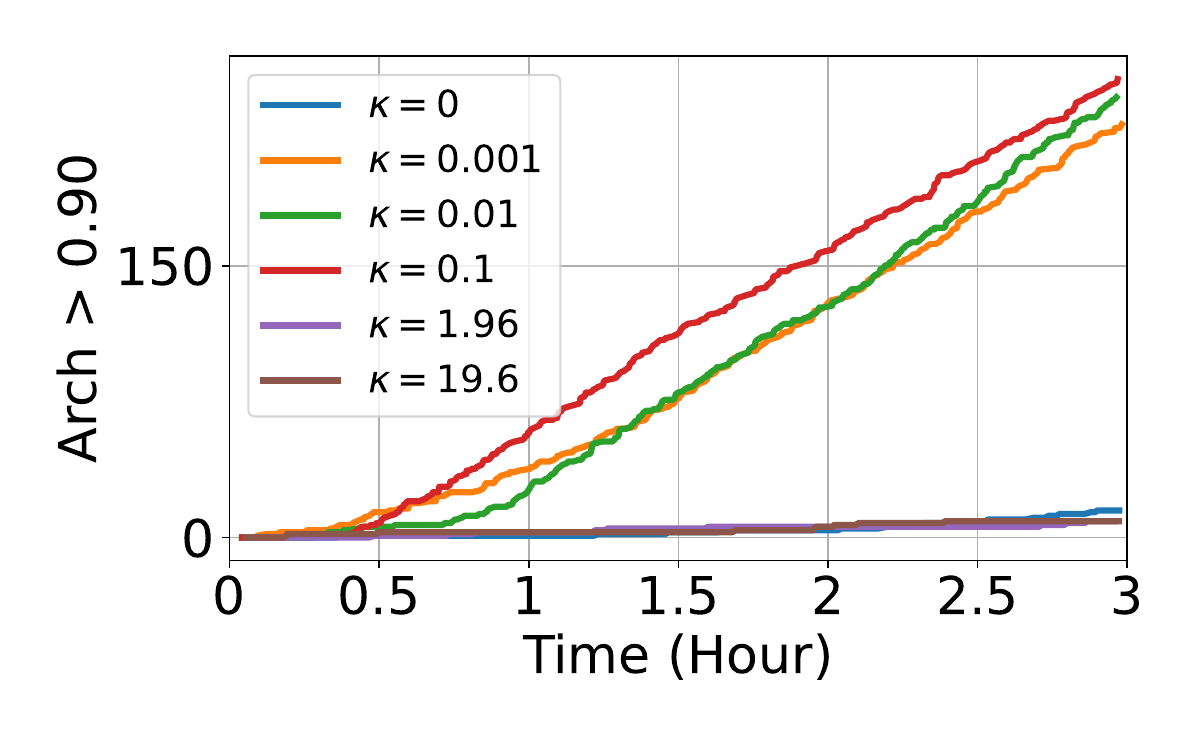}
        \caption{}
        \label{fig:agebo-exploration-unique-architectures}
    \end{subfigure}
\vspace{-0.3cm}
    \caption{Comparison of AgEBO-4-8 with different $\kappa$ values (the higher $\kappa$ the more exploration). (a) is the evolution of the best objective. (b) is the number of unique architectures better than the baseline. \vspace{-0.3cm}}
    \label{fig:agebo-exploration}
\end{figure}

\edit{Figure \ref{fig:agebo-exploration-objective} shows the best validation $R^2$ found by the search over time. Pure exploitation ($\kappa = 0$), balance between exploration and exploitation (default $\kappa = 1.96$) large exploration ($\kappa = 19.6$) did not result in high accuracy.  Figure~\ref{fig:agebo-exploration-unique-architectures} shows the number of architectures performing better than the baseline found by AgEBO for  different $\kappa$ values. We can  observe that, AgEBO with the $\kappa$ value of \{0.001, 0.01, 0.1\} (stronger exploitation) outperforms those with 1.96 (balance between exploration and exploitation) and 19.6 (stronger exploration) with respect to the number of architectures performing better than the baseline (by two orders of magnitude) and the time needed to reach a better solution shown in Figure~\ref{fig:agebo-exploration-objective}.
The exploration of hyperparameter values in AgEBO with $\kappa$ value of \{0.001, 0.01, 0.1\} happens only in the random initialization phase. During the iterative phase, given the stronger exploitation setting, hyperparameter configurations are generated close to the best ones found so far in the search. On the other hand, there is a significant degree of exploration with $\kappa$ values of 1.96 and 19.6. This results in a lot more low performing configurations of hyperparameters.}

\edit{As discussed in Section \ref{mixed-age}, the AgEBO's effectiveness is attributed to the implicit regularization mechanism of AgE method and a stronger exploitation of in the BO method. When the BO method have stronger exploration, it generates hyperparameter configurations that are different from the previously best performing hyperparameter values. When these values are used for the high-performing architectures after inheritance and retraining, they will most likely become low performing ones and removed from the population. Therefore BO should generate hyperparameter values that are close to the previously found ones. However, it should be noted that a pure exploitation  ($\kappa = 0$) will keep generating the same hyperparameter values after a few generations, which did not help AgEBO to generate high-performing architectures.} 

%\edit{For this study, it is important to keep in mind the relatively low number of 16 parallel workers where each worker is using 4 GPUs. Exploration is useful to exit already exploited local minimas and therefore requires a lot more evaluations. In fact, pure exploration is similar to random search which become more and more efficient when the number of computational resources increase. Therefore the $\kappa$-value can be increased when the number of evaluations is significantly bigger. }

\subsection{Synchronous vs Asynchronous AgEBO}

Compared to synchronous BO, asynchronous methods have received relatively less attention in the BO community. Therefore, we compare synchronous and  asynchronous BO with constant liar strategy in the AgEBO to show that the latter is effective.

The synchronous version of AgEBO is obtained by placing a synchronisation barrier after line 9  ( \texttt{get\_finished\_evaluations()}) in Algorithm~\ref{alg:AgEBO}. Specifically, AgEBO proceeds only after getting the evaluation results.  
For this experiment, we used AgEBO-2-8 (2 GPUs per evaluation and 8 nodes) on the Combo data set. %which is the best setting according to Table \ref{tab:best-networks-attn} and \ref{tab:best-networks-combo}. 
The synchronous and asynchronous versions are referred to as AgEBO-2-8 sync and AgEBO-2-8 async, respectively.

\begin{figure}[t]
        \includegraphics[width=.45\textwidth]{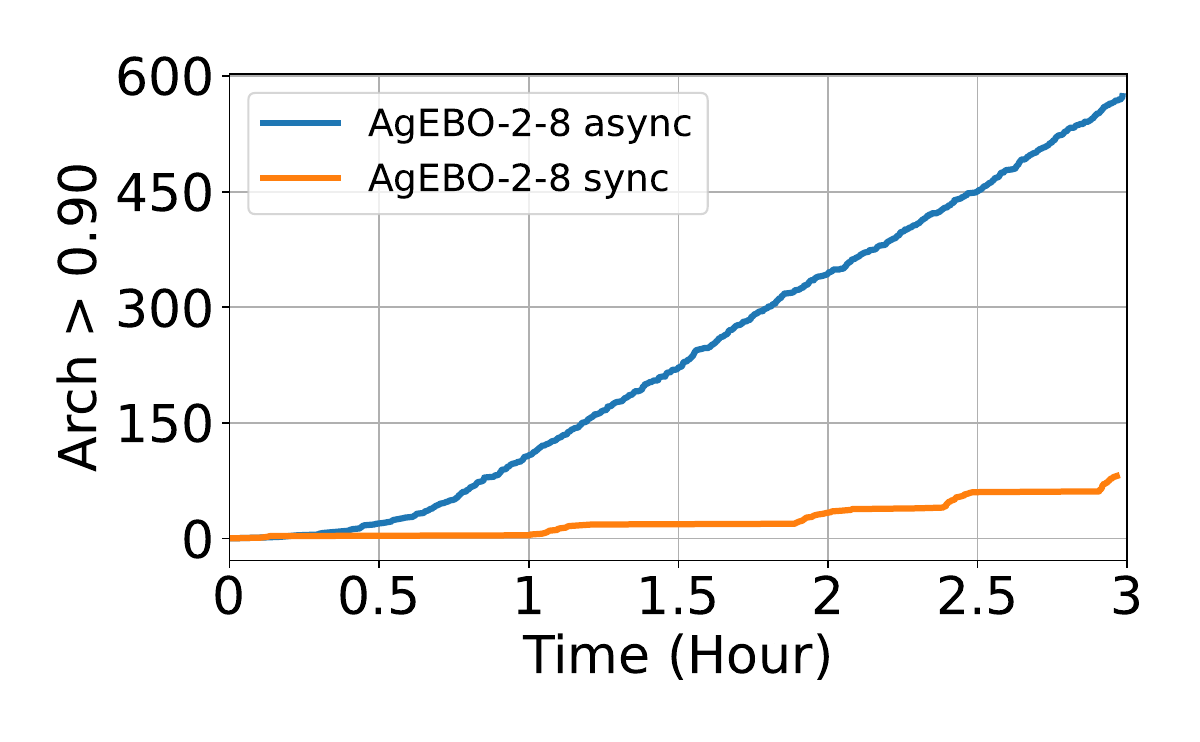}
        \vspace{-0.5cm}
    \caption{ Comparison of synchronous and asynchronous AgEBO on the Combo data set. The number of unique architectures found by the search methods that are better than the baseline. \vspace{-0.8cm}}
    \label{fig:agebo-sync-async}
\end{figure}

Figure~\ref{fig:agebo-sync-async} shows the number of unique architectures found by AgEBO-2-8 sync and AgEBO-2-8 async that are better than the baseline, respectively. These results show that the asynchronous approach is better than the synchronous one. 
Both synchronous and asynchronous outperforms baseline in about 6 mins. 
However, we observed that AgEBO-2-8 async has a much better node utilization 95\% whereas it was only 26\% for AgEBO-2-8 async.  Consequently, the number of evaluations performed on AgEBO-2-8 async is 4.77x more than AgEBO-2-8 sync (759 for the former, 159 for the latter). 
The best $R^2$ value found by AgEBO-2-8 async was 0.936 whereas it was 0.926 for AgEBO-2-8 sync. While a few sophisticated asynchronous BO methods exist in the literature \cite{kandasamy2019tuning,pmlr-v97-alvi19a,balandat2019botorch}, they seek to achieve balance between exploration and exploitation. Nevertheless, such methods are not required in our setting because of the synergy between AgE and BO with stronger exploitation.

%

% combo_2gpu_8_agebo  ->  0.936 R2
% combo_2gpu_8_agebo_sync  ->  0.926 R@
% AgEBO-2-8 async  ->  759 evaluations
% AgEBO-2-8 sync  ->  152 evaluations
% Usage: AgEBO-2-8 async -> 94.7167622839549%
% Usage: AgEBO-2-8 sync -> 26.21858711740761%
% combo_2gpu_8_agebo  ->  7.19 ± 4.39
% combo_2gpu_8_agebo_sync  ->  9.94 ± 8.54
% combo_2gpu_8_agebo  ->  6.15 min to 0.902
% combo_2gpu_8_agebo_sync  ->  55.56 min to 0.902
% Training time best arch (AgEBO-2-8 async): 4.29 min. and 0.09 min/epoch
% Training time best arch (AgEBO-2-8 sync): 8.65 min. and 0.09 min/epoch

\begin{comment}
\begin{figure}[!h]
    \centering
    \begin{subfigure}{\linewidth}
        \centering
        \includegraphics[width=.8\textwidth]{plot_objective_multi_sync_async.pdf}
        \caption{}
        \label{fig:agebo-sync-async-objective}
    \end{subfigure}
    \begin{subfigure}{\linewidth}
        \centering
        \includegraphics[width=.8\textwidth]{plot_count_arch_better_than_baseline_sync_async.pdf}
        \caption{}
        \label{fig:agebo-sync-async-unique-architectures}
    \end{subfigure}

    \caption{Comparison of synchronous and asynchronous AgEBO on the Combo data set. The number of unique architectures found by the search methods that are better than the baseline.}
    \label{fig:agebo-sync-async}
\end{figure}
\end{comment}

\section{Related Work}

From the novelty perspective, our method has three components: hyperparameter search for data-parallel training, joint NAS and HPS, and application to tabular data. We review the related work from the perspective of each component.

%and highlight our contributions. 

The literature on HPS for tuning hyperparameters on  distributed data-parallel training to optimize learning rate, batch size, and  others is  limited. A commonly used approach to adapt learning rate and batch size in distributed data-parallel training is the linear scaling rule. The values of the learning rate and batch size used for the single-process training are multiplied by the number of processes in distributed data-parallel training. 
In an Amazon blog \cite{amazonBlogPostDataParallelTuning}, the importance of tuning learning rate and batch size for a given number of GPUs in data-parallel training has been discussed. Specifically, the Amazon SageMaker HPO tool has been used as a proof of concept; but the study was not performed at scale, and the effectiveness was not assessed on wide range of data sets. The use of BO to tune the learning rate, batch size, and  other hyperparameters while using distributed training has never been investigated before.

Within NAS, several approaches have been proposed to reduce the training time. Examples include using smaller architecture search and stacking ~\cite{zoph_neural_2016,zoph_learning_2017}, reducing the number of epochs~\cite{zela_towards_2018}, computing the validation performance from a randomly initialised DNN~\cite{zela_nas-bench-1shot1_2020}, estimating the accuracy performance of DNN for a large budget (time) when trained with a smaller budget~\cite{zheng_rethinking_2020}, sharing the weights of previously trained DNN~\cite{pham_efficient_2018}, imposing a time budget~\cite{balaprakash_scalable_2019}, and using information from data relatively to an initialised DNN (but only for convolution NN) without training~\cite{mellor_neural_2020}. 
These methods have several limitations. 
Stacking the simpler model is feasible for image data sets but can lead to overfitting in tabular data sets; and reducing the epochs and time budget during NAS can 
%lead to poor relative ranking between the small and extensive budget and eventually 
result in a low performing model~\cite{zela_towards_2018}.   
Compared with these methods, data-parallel training is a promising approach because of its ability to match with the learning curve of the classical training while consequently speeding up the training~\cite{goyal_accurate_2018}. Nevertheless, the use of data-parallel training within NAS was not investigated before.

The joint NAS and HPS approach that we propose is similar to BO Hyperband (BOHB)~\cite{zela_towards_2018}. 
It considers the joint space and uses a multivariate kernel density estimation model to sample promising configurations. The sampled configurations are evaluated by using a successive
halving approach, where promising configurations are allowed to run longer with more resources. 
%Our approach differs from BOHB in the following ways. 
Compared to our approach, BOHB does not 1) differentiate the model hyperparameters from neural architecture hyperparameters; and 2) utilize data-parallel training to speedup the evaluation of neural networks, but instead adopt successive halving to allocate more resources to promising configurations. 
%This is a blocking approach, 
While it can be quite effective under limited resource setting, scaling the successive halving method can lead to poor node utilization.
%due to blocking and synchronisation.

AutoML for tabular data has received considerable attention in recent years. Notable examples include  auto-sklearn~\cite{feurer2020autosklearn}, Auto-WEKA~\cite{thornton_auto-weka_2013}, 
H2O AutoML~\cite{H2OAutoML}, and TPOPT~\cite{OlsonGECCO2016}. A benchmark~\cite{amlb2019} of these methods was conducted to compare their performance on different data sets. The auto-sklearn approach proved more robust in general. We tested auto-sklearn on our data sets but the results were poor.  Recently,  AutoGluon~\cite{erickson_autogluon-tabular_2020} and Auto-PyTorch~\cite{zimmer_auto-pytorch_2020} have emerged as new AutoML methods for tabular data. AutoGluon  uses an ensemble of many different learning algorithms to boost their performance. Auto-PyTorch  also uses an ensemble approach, but the models are restricted to DNNs. As an exploratory study, we used four large tabular data sets from the OpenML 
 benchmark~\cite{feurer-arxiv19a}. The selection was motivated by a tabular data
 benchmark study using AutoGluon. Among all
 the data sets benchmarked with AutoGluon, we selected the four largest data sets (Covertype~\cite{ucikddarchive}:, Airlines~\cite{airlinesdataset}, Albert~\cite{automlchallenges}, and Dionis~\cite{automlchallenges}) having the largest number of data points. We found that the prediction accuracy of AgEBO is better than or comparable to that of AutoGluon and Auto-PyTorch and provides a significant advantage with respect to the inference time.

\section{Conclusion and Future Work}

We developed AgEBO-Tabular, a joint neural architecture (NAS) and hyperparameter search (HPS) method to discover high-performing neural network models for tabular data. We developed an architecture search space for generating fully connected neural networks with skip connections and possibly with multiple inputs. The search method combines two distinct methods: (1) aging evolution (AgE), a parallel NAS to search over the architecture decision variables; and (2) an asynchronous Bayesian optimization (BO) method to automatically tune the hyperparameters of data-parallel training in order to reduce the evaluation time of each architecture. 

We showed that using data-parallel training in AgE without tuning the learning rate, batch size, and other hyperparameters can adversely affect the accuracy. Then, we demonstrated that AgEBO can improve the accuracy of the discovered models and the time to generate high-performing neural networks. We compared the best-discovered models from AgEBO with state-of-the-art human-made models and showed that AgEBO outperforms the state-of-the-art human made models both in accuracy and training time.

The analysis of the best values obtained by AgEBO showed the need for data-set-specific tuning. Moreover, we showed that, unlike the typical BO that balances exploration and exploitation, a stronger exploitation is critical for AgEBO for generating high-performing models in short computation time.

The algorithmic components (AgE and BO) of the proposed AgEBO method are not specific to NVIDIA accelerator model. We can generalize and adapt AgEBO on CPU based clusters or other CPU+accelerator type platforms. Moreover, the only component that is specific to the tabular data set is the search space. We will expand the search space for image, text, and graph data and evaluate the effectiveness of the proposed AgEBO method.

Our other future work will include (1) applying AgEBO to generate neural architectures for  other data types such as images, texts, and graphs; (2) developing multinode data-parallel training within NAS for larger data sets; (3) developing meta-learning and transfer learning approaches to reuse the knowledge and results from previous experimental runs for related data sets; and \edit{(4) adaptive and flexible workflow and scheduler system for joint model and resource optimization.}

\section*{Acknowledgment}
This material is based upon work supported by the U.S.\ Department of Energy 
(DOE), Office of Science, Office of Advanced Scientific Computing Research, under
Contract DE-AC02-06CH11357. This research used resources of the Argonne 
Leadership Computing Facility, which is a DOE Office of Science User Facility.

\bibliographystyle{ACM-Reference-Format}
\bibliography{others,references,appendix}


\begin{thebibliography}{41}

%%% the Web address.

\ifx \showCODEN    \undefined \def \showCODEN     #1{\unskip}     \fi
\ifx \showDOI      \undefined \def \showDOI       #1{#1}\fi
\ifx \showISBNx    \undefined \def \showISBNx     #1{\unskip}     \fi
\ifx \showISBNxiii \undefined \def \showISBNxiii  #1{\unskip}     \fi
\ifx \showISSN     \undefined \def \showISSN      #1{\unskip}     \fi
\ifx \showLCCN     \undefined \def \showLCCN      #1{\unskip}     \fi
\ifx \shownote     \undefined \def \shownote      #1{#1}          \fi
\ifx \showarticletitle \undefined \def \showarticletitle #1{#1}   \fi
\ifx \showURL      \undefined \def \showURL       {\relax}        \fi
% invisible to TeX
\providecommand\bibfield[2]{#2}
\providecommand\bibinfo[2]{#2}
\providecommand\natexlab[1]{#1}
\providecommand\showeprint[2][]{arXiv:#2}

\bibitem[\protect\citeauthoryear{??}{ama}{[n.d.]}]%
        {amazonBlogPostDataParallelTuning}
 \bibinfo{year}{[n.d.]}\natexlab{}.
\newblock \bibinfo{title}{{The importance of hyperparameter tuning for scaling
  deep learning training to multiple GPUs}, howpublished =
  {https://aws.amazon.com/blogs/machine-learning/the-importance-of-hyperparameter-tuning-for-scaling-deep-learning-training-to-multiple-gpus/},
  note = {Accessed: 2020-10-08}}.
\newblock
\newblock


\bibitem[\protect\citeauthoryear{Albert~Bifet}{Albert~Bifet}{2009}]%
        {airlinesdataset}
\bibfield{author}{\bibinfo{person}{Elena~Ikonomovska Albert~Bifet}.}
  \bibinfo{year}{2009}\natexlab{}.
\newblock \bibinfo{booktitle}{\emph{Airlines Dataset Inspired in the regression
  dataset from Elena Ikonomovska. The task is to predict whether a given flight
  will be delayed, given the information of the scheduled departure.}}
\newblock
\urldef\tempurl%
\url{http://kt.ijs.si/elena\_ikonomovska/data.html}
\showURL{%
\tempurl}


\bibitem[\protect\citeauthoryear{Alvi, Ru, Calliess, Roberts, and Osborne}{Alvi
  et~al\mbox{.}}{2019}]%
        {pmlr-v97-alvi19a}
\bibfield{author}{\bibinfo{person}{Ahsan Alvi}, \bibinfo{person}{Binxin Ru},
  \bibinfo{person}{Jan-Peter Calliess}, \bibinfo{person}{Stephen Roberts},
  {and} \bibinfo{person}{Michael~A. Osborne}.} \bibinfo{year}{2019}\natexlab{}.
\newblock \showarticletitle{Asynchronous Batch {B}ayesian Optimisation with
  Improved Local Penalisation}. In \bibinfo{booktitle}{\emph{Proceedings of the
  36th International Conference on Machine Learning}}
  \emph{(\bibinfo{series}{Proceedings of Machine Learning Research},
  Vol.~\bibinfo{volume}{97})}, \bibfield{editor}{\bibinfo{person}{Kamalika
  Chaudhuri} {and} \bibinfo{person}{Ruslan Salakhutdinov}} (Eds.).
  \bibinfo{publisher}{PMLR}, \bibinfo{pages}{253--262}.
\newblock
\urldef\tempurl%
\url{http://proceedings.mlr.press/v97/alvi19a.html}
\showURL{%
\tempurl}


\bibitem[\protect\citeauthoryear{Balandat, Karrer, Jiang, Daulton, Letham,
  Wilson, and Bakshy}{Balandat et~al\mbox{.}}{2019}]%
        {balandat2019botorch}
\bibfield{author}{\bibinfo{person}{Maximilian Balandat}, \bibinfo{person}{Brian
  Karrer}, \bibinfo{person}{Daniel~R Jiang}, \bibinfo{person}{Samuel Daulton},
  \bibinfo{person}{Benjamin Letham}, \bibinfo{person}{Andrew~Gordon Wilson},
  {and} \bibinfo{person}{Eytan Bakshy}.} \bibinfo{year}{2019}\natexlab{}.
\newblock \showarticletitle{BoTorch: A Framework for Efficient Monte-Carlo
  Bayesian Optimization}.
\newblock \bibinfo{journal}{\emph{arXiv preprint arXiv:1910.06403}}
  (\bibinfo{year}{2019}).
\newblock


\bibitem[\protect\citeauthoryear{Balaprakash, Egele, Salim, Vishwanath, Wild,
  Jha, Dorier, Felker, Maulik, and Lusch}{Balaprakash et~al\mbox{.}}{2020}]%
        {deephyper_software}
\bibfield{author}{\bibinfo{person}{Prasanna Balaprakash},
  \bibinfo{person}{Romain Egele}, \bibinfo{person}{Michael Salim},
  \bibinfo{person}{Venkat Vishwanath}, \bibinfo{person}{Stefan Wild},
  \bibinfo{person}{Dipendra Jha}, \bibinfo{person}{Matthieu Dorier},
  \bibinfo{person}{Kyle~Gerard Felker}, \bibinfo{person}{Romit Maulik}, {and}
  \bibinfo{person}{Bethany Lusch}.} \bibinfo{year}{2020}\natexlab{}.
\newblock \bibinfo{booktitle}{\emph{deephyper/deephyper: 0.1.12}}.
\newblock
\urldef\tempurl%
\url{https://github.com/deephyper/deephyper}
\showURL{%
\tempurl}


\bibitem[\protect\citeauthoryear{Balaprakash, Egele, Salim, Wild, Vishwanath,
  Xia, Brettin, and Stevens}{Balaprakash et~al\mbox{.}}{[n.d.]}]%
        {balaprakash_scalable_2019}
\bibfield{author}{\bibinfo{person}{Prasanna Balaprakash},
  \bibinfo{person}{Romain Egele}, \bibinfo{person}{Misha Salim},
  \bibinfo{person}{Stefan Wild}, \bibinfo{person}{Venkatram Vishwanath},
  \bibinfo{person}{Fangfang Xia}, \bibinfo{person}{Tom Brettin}, {and}
  \bibinfo{person}{Rick Stevens}.} \bibinfo{year}{[n.d.]}\natexlab{}.
\newblock \showarticletitle{Scalable Reinforcement-Learning-Based Neural
  Architecture Search for Cancer Deep Learning Research}.
\newblock  (\bibinfo{year}{[n.\,d.]}), \bibinfo{pages}{1--33}.
\newblock
\urldef\tempurl%
\url{https://doi.org/10.1145/3295500.3356202}
\showDOI{\tempurl}
\showeprint[arxiv]{1909.00311}


\bibitem[\protect\citeauthoryear{Chevalier and Ginsbourger}{Chevalier and
  Ginsbourger}{2013}]%
        {chevalier2013fast}
\bibfield{author}{\bibinfo{person}{Cl{\'e}ment Chevalier} {and}
  \bibinfo{person}{David Ginsbourger}.} \bibinfo{year}{2013}\natexlab{}.
\newblock \showarticletitle{Fast computation of the multi-points expected
  improvement with applications in batch selection}. In
  \bibinfo{booktitle}{\emph{International Conference on Learning and
  Intelligent Optimization}}. Springer, \bibinfo{pages}{59--69}.
\newblock


\bibitem[\protect\citeauthoryear{Chu, Zhou, Zhang, and Li}{Chu
  et~al\mbox{.}}{[n.d.]}]%
        {chu_fair_2020}
\bibfield{author}{\bibinfo{person}{Xiangxiang Chu}, \bibinfo{person}{Tianbao
  Zhou}, \bibinfo{person}{Bo Zhang}, {and} \bibinfo{person}{Jixiang Li}.}
  \bibinfo{year}{[n.d.]}\natexlab{}.
\newblock \showarticletitle{Fair {DARTS}: Eliminating Unfair Advantages in
  Differentiable Architecture Search}.
\newblock  (\bibinfo{year}{[n.\,d.]}).
\newblock
\showeprint[arxiv]{1911.12126}
\urldef\tempurl%
\url{http://arxiv.org/abs/1911.12126}
\showURL{%
\tempurl}


\bibitem[\protect\citeauthoryear{Clyde, Brettin, Partin, Shaulik, Yoo, Evrard,
  Zhu, Xia, and Stevens}{Clyde et~al\mbox{.}}{2020}]%
        {clyde2020systematic}
\bibfield{author}{\bibinfo{person}{Austin Clyde}, \bibinfo{person}{Tom
  Brettin}, \bibinfo{person}{Alexander Partin}, \bibinfo{person}{Maulik
  Shaulik}, \bibinfo{person}{Hyunseung Yoo}, \bibinfo{person}{Yvonne Evrard},
  \bibinfo{person}{Yitan Zhu}, \bibinfo{person}{Fangfang Xia}, {and}
  \bibinfo{person}{Rick Stevens}.} \bibinfo{year}{2020}\natexlab{}.
\newblock \showarticletitle{A Systematic Approach to Featurization for Cancer
  Drug Sensitivity Predictions with Deep Learning}.
\newblock \bibinfo{journal}{\emph{arXiv preprint arXiv:2005.00095}}
  (\bibinfo{year}{2020}).
\newblock


\bibitem[\protect\citeauthoryear{Erickson, Mueller, Shirkov, Zhang, Larroy, Li,
  and Smola}{Erickson et~al\mbox{.}}{[n.d.]}]%
        {erickson_autogluon-tabular_2020}
\bibfield{author}{\bibinfo{person}{Nick Erickson}, \bibinfo{person}{Jonas
  Mueller}, \bibinfo{person}{Alexander Shirkov}, \bibinfo{person}{Hang Zhang},
  \bibinfo{person}{Pedro Larroy}, \bibinfo{person}{Mu Li}, {and}
  \bibinfo{person}{Alexander Smola}.} \bibinfo{year}{[n.d.]}\natexlab{}.
\newblock \showarticletitle{{AutoGluon}-Tabular: Robust and Accurate {AutoML}
  for Structured Data}.
\newblock  (\bibinfo{year}{[n.\,d.]}).
\newblock
\showeprint[arxiv]{2003.06505}
\urldef\tempurl%
\url{http://arxiv.org/abs/2003.06505}
\showURL{%
\tempurl}


\bibitem[\protect\citeauthoryear{Fern{\'a}ndez-Delgado, Cernadas, Barro, and
  Amorim}{Fern{\'a}ndez-Delgado et~al\mbox{.}}{2014}]%
        {fernandez2014we}
\bibfield{author}{\bibinfo{person}{Manuel Fern{\'a}ndez-Delgado},
  \bibinfo{person}{Eva Cernadas}, \bibinfo{person}{Sen{\'e}n Barro}, {and}
  \bibinfo{person}{Dinani Amorim}.} \bibinfo{year}{2014}\natexlab{}.
\newblock \showarticletitle{Do we need hundreds of classifiers to solve real
  world classification problems?}
\newblock \bibinfo{journal}{\emph{The journal of machine learning research}}
  \bibinfo{volume}{15}, \bibinfo{number}{1} (\bibinfo{year}{2014}),
  \bibinfo{pages}{3133--3181}.
\newblock


\bibitem[\protect\citeauthoryear{Feurer, Eggensperger, Falkner, Lindauer, and
  Hutter}{Feurer et~al\mbox{.}}{2020}]%
        {feurer2020autosklearn}
\bibfield{author}{\bibinfo{person}{Matthias Feurer}, \bibinfo{person}{Katharina
  Eggensperger}, \bibinfo{person}{Stefan Falkner}, \bibinfo{person}{Marius
  Lindauer}, {and} \bibinfo{person}{Frank Hutter}.}
  \bibinfo{year}{2020}\natexlab{}.
\newblock \bibinfo{title}{Auto-Sklearn 2.0: The Next Generation}.
\newblock
\newblock
\showeprint[arxiv]{2007.04074}~[cs.LG]


\bibitem[\protect\citeauthoryear{Feurer, van Rijn, Kadra, Gijsbers, Mallik,
  Ravi, Müller, Vanschoren, and Hutter}{Feurer et~al\mbox{.}}{2019}]%
        {feurer-arxiv19a}
\bibfield{author}{\bibinfo{person}{Matthias Feurer}, \bibinfo{person}{Jan~N.
  van Rijn}, \bibinfo{person}{Arlind Kadra}, \bibinfo{person}{Pieter Gijsbers},
  \bibinfo{person}{Neeratyoy Mallik}, \bibinfo{person}{Sahithya Ravi},
  \bibinfo{person}{Andreas Müller}, \bibinfo{person}{Joaquin Vanschoren},
  {and} \bibinfo{person}{Frank Hutter}.} \bibinfo{year}{2019}\natexlab{}.
\newblock \showarticletitle{OpenML-Python: an extensible Python API for
  OpenML}.
\newblock \bibinfo{journal}{\emph{arXiv:1911.02490}} (\bibinfo{year}{2019}).
\newblock


\bibitem[\protect\citeauthoryear{Gijsbers, LeDell, Poirier, Thomas, Bischl, and
  Vanschoren}{Gijsbers et~al\mbox{.}}{2019}]%
        {amlb2019}
\bibfield{author}{\bibinfo{person}{P. Gijsbers}, \bibinfo{person}{E. LeDell},
  \bibinfo{person}{S. Poirier}, \bibinfo{person}{J. Thomas},
  \bibinfo{person}{B. Bischl}, {and} \bibinfo{person}{J. Vanschoren}.}
  \bibinfo{year}{2019}\natexlab{}.
\newblock \showarticletitle{An Open Source AutoML Benchmark}.
\newblock \bibinfo{journal}{\emph{arXiv preprint arXiv:1907.00909 [cs.LG]}}
  (\bibinfo{year}{2019}).
\newblock
\urldef\tempurl%
\url{https://arxiv.org/abs/1907.00909}
\showURL{%
\tempurl}
\newblock
\shownote{Accepted at AutoML Workshop at ICML 2019.}


\bibitem[\protect\citeauthoryear{Ginsbourger, Le~Riche, and
  Carraro}{Ginsbourger et~al\mbox{.}}{[n.d.]}]%
        {hiot_kriging_2010}
\bibfield{author}{\bibinfo{person}{David Ginsbourger},
  \bibinfo{person}{Rodolphe Le~Riche}, {and} \bibinfo{person}{Laurent
  Carraro}.} \bibinfo{year}{[n.d.]}\natexlab{}.
\newblock \showarticletitle{Kriging Is Well-Suited to Parallelize
  Optimization}.
\newblock In \bibinfo{booktitle}{\emph{Computational Intelligence in Expensive
  Optimization Problems}}, \bibfield{editor}{\bibinfo{person}{Yoel Tenne} {and}
  \bibinfo{person}{Chi-Keong Goh}} (Eds.). Vol.~\bibinfo{volume}{2}.
  \bibinfo{publisher}{Springer Berlin Heidelberg}, \bibinfo{pages}{131--162}.
\newblock
\showISBNx{978-3-642-10700-9 978-3-642-10701-6}
\urldef\tempurl%
\url{https://doi.org/10.1007/978-3-642-10701-6_6}
\showDOI{\tempurl}
\newblock
\shownote{Series Title: Adaptation Learning and Optimization.}


\bibitem[\protect\citeauthoryear{Goyal, Doll{\'{a}}r, Girshick, Noordhuis,
  Wesolowski, Kyrola, Tulloch, Jia, and He}{Goyal et~al\mbox{.}}{2017}]%
        {DBLP:journals/corr/GoyalDGNWKTJH17}
\bibfield{author}{\bibinfo{person}{Priya Goyal}, \bibinfo{person}{Piotr
  Doll{\'{a}}r}, \bibinfo{person}{Ross~B. Girshick}, \bibinfo{person}{Pieter
  Noordhuis}, \bibinfo{person}{Lukasz Wesolowski}, \bibinfo{person}{Aapo
  Kyrola}, \bibinfo{person}{Andrew Tulloch}, \bibinfo{person}{Yangqing Jia},
  {and} \bibinfo{person}{Kaiming He}.} \bibinfo{year}{2017}\natexlab{}.
\newblock \showarticletitle{Accurate, Large Minibatch {SGD:} Training ImageNet
  in 1 Hour}.
\newblock \bibinfo{journal}{\emph{CoRR}}  \bibinfo{volume}{abs/1706.02677}
  (\bibinfo{year}{2017}).
\newblock
\showeprint[arxiv]{1706.02677}
\urldef\tempurl%
\url{http://arxiv.org/abs/1706.02677}
\showURL{%
\tempurl}


\bibitem[\protect\citeauthoryear{Goyal, Dollár, Girshick, Noordhuis,
  Wesolowski, Kyrola, Tulloch, Jia, and He}{Goyal et~al\mbox{.}}{[n.d.]}]%
        {goyal_accurate_2018}
\bibfield{author}{\bibinfo{person}{Priya Goyal}, \bibinfo{person}{Piotr
  Dollár}, \bibinfo{person}{Ross Girshick}, \bibinfo{person}{Pieter
  Noordhuis}, \bibinfo{person}{Lukasz Wesolowski}, \bibinfo{person}{Aapo
  Kyrola}, \bibinfo{person}{Andrew Tulloch}, \bibinfo{person}{Yangqing Jia},
  {and} \bibinfo{person}{Kaiming He}.} \bibinfo{year}{[n.d.]}\natexlab{}.
\newblock \showarticletitle{Accurate, Large Minibatch {SGD}: Training
  {ImageNet} in 1 Hour}.
\newblock  (\bibinfo{year}{[n.\,d.]}).
\newblock
\showeprint[arxiv]{1706.02677}
\urldef\tempurl%
\url{http://arxiv.org/abs/1706.02677}
\showURL{%
\tempurl}


\bibitem[\protect\citeauthoryear{Guyon, Sun-Hosoya, Boull\'e, Escalante,
  Escalera, Liu, Jajetic, Ray, Saeed, Sebag, Statnikov, Tu, and Viegas}{Guyon
  et~al\mbox{.}}{2019}]%
        {automlchallenges}
\bibfield{author}{\bibinfo{person}{Isabelle Guyon}, \bibinfo{person}{Lisheng
  Sun-Hosoya}, \bibinfo{person}{Marc Boull\'e}, \bibinfo{person}{Hugo~Jair
  Escalante}, \bibinfo{person}{Sergio Escalera}, \bibinfo{person}{Zhengying
  Liu}, \bibinfo{person}{Damir Jajetic}, \bibinfo{person}{Bisakha Ray},
  \bibinfo{person}{Mehreen Saeed}, \bibinfo{person}{Mich\'ele Sebag},
  \bibinfo{person}{Alexander Statnikov}, \bibinfo{person}{WeiWei Tu}, {and}
  \bibinfo{person}{Evelyne Viegas}.} \bibinfo{year}{2019}\natexlab{}.
\newblock \showarticletitle{Analysis of the AutoML Challenge series 2015-2018}.
  In \bibinfo{booktitle}{\emph{AutoML}} \emph{(\bibinfo{series}{Springer series
  on Challenges in Machine Learning})}.
\newblock
\urldef\tempurl%
\url{https://www.automl.org/wp-content/uploads/2018/09/chapter10-challenge.pdf}
\showURL{%
\tempurl}


\bibitem[\protect\citeauthoryear{H2O.ai}{H2O.ai}{2017}]%
        {H2OAutoML}
\bibfield{author}{\bibinfo{person}{H2O.ai}.} \bibinfo{year}{2017}\natexlab{}.
\newblock \bibinfo{booktitle}{\emph{H2O AutoML}}.
\newblock
\urldef\tempurl%
\url{http://docs.h2o.ai/h2o/latest-stable/h2o-docs/automl.html}
\showURL{%
\tempurl}
\newblock
\shownote{H2O version 3.30.0.1.}


\bibitem[\protect\citeauthoryear{Hettich and Bay}{Hettich and Bay}{1999}]%
        {ucikddarchive}
\bibfield{author}{\bibinfo{person}{S. Hettich} {and} \bibinfo{person}{S.~D.
  Bay}.} \bibinfo{year}{1999}\natexlab{}.
\newblock \bibinfo{booktitle}{\emph{The UCI KDD Archive}}.
\newblock
\urldef\tempurl%
\url{http://kdd.ics.uci.edu}
\showURL{%
\tempurl}


\bibitem[\protect\citeauthoryear{Kandasamy, Vysyaraju, Neiswanger, Paria,
  Collins, Schneider, Poczos, and Xing}{Kandasamy et~al\mbox{.}}{2019}]%
        {kandasamy2019tuning}
\bibfield{author}{\bibinfo{person}{Kirthevasan Kandasamy},
  \bibinfo{person}{Karun~Raju Vysyaraju}, \bibinfo{person}{Willie Neiswanger},
  \bibinfo{person}{Biswajit Paria}, \bibinfo{person}{Christopher~R Collins},
  \bibinfo{person}{Jeff Schneider}, \bibinfo{person}{Barnabas Poczos}, {and}
  \bibinfo{person}{Eric~P Xing}.} \bibinfo{year}{2019}\natexlab{}.
\newblock \showarticletitle{Tuning hyperparameters without grad students:
  Scalable and robust bayesian optimisation with dragonfly}.
\newblock \bibinfo{journal}{\emph{arXiv preprint arXiv:1903.06694}}
  (\bibinfo{year}{2019}).
\newblock


\bibitem[\protect\citeauthoryear{Kingma and Ba}{Kingma and Ba}{2014}]%
        {kingma2014adam}
\bibfield{author}{\bibinfo{person}{Diederik~P Kingma} {and}
  \bibinfo{person}{Jimmy Ba}.} \bibinfo{year}{2014}\natexlab{}.
\newblock \showarticletitle{Adam: A method for stochastic optimization}.
\newblock \bibinfo{journal}{\emph{arXiv preprint arXiv:1412.6980}}
  (\bibinfo{year}{2014}).
\newblock


\bibitem[\protect\citeauthoryear{Mellor, Turner, Storkey, and Crowley}{Mellor
  et~al\mbox{.}}{[n.d.]}]%
        {mellor_neural_2020}
\bibfield{author}{\bibinfo{person}{Joseph Mellor}, \bibinfo{person}{Jack
  Turner}, \bibinfo{person}{Amos Storkey}, {and} \bibinfo{person}{Elliot~J.
  Crowley}.} \bibinfo{year}{[n.d.]}\natexlab{}.
\newblock \showarticletitle{Neural Architecture Search without Training}.
\newblock  (\bibinfo{year}{[n.\,d.]}).
\newblock
\showeprint[arxiv]{2006.04647}
\urldef\tempurl%
\url{http://arxiv.org/abs/2006.04647}
\showURL{%
\tempurl}


\bibitem[\protect\citeauthoryear{Moritz, Nishihara, Wang, Tumanov, Liaw, Liang,
  Elibol, Yang, Paul, Jordan, and Stoica}{Moritz et~al\mbox{.}}{2018}]%
        {moritz2018ray}
\bibfield{author}{\bibinfo{person}{Philipp Moritz}, \bibinfo{person}{Robert
  Nishihara}, \bibinfo{person}{Stephanie Wang}, \bibinfo{person}{Alexey
  Tumanov}, \bibinfo{person}{Richard Liaw}, \bibinfo{person}{Eric Liang},
  \bibinfo{person}{Melih Elibol}, \bibinfo{person}{Zongheng Yang},
  \bibinfo{person}{William Paul}, \bibinfo{person}{Michael~I. Jordan}, {and}
  \bibinfo{person}{Ion Stoica}.} \bibinfo{year}{2018}\natexlab{}.
\newblock \bibinfo{title}{Ray: A Distributed Framework for Emerging AI
  Applications}.
\newblock
\newblock
\showeprint[arxiv]{1712.05889}~[cs.DC]


\bibitem[\protect\citeauthoryear{Olson, Bartley, Urbanowicz, and Moore}{Olson
  et~al\mbox{.}}{2016}]%
        {OlsonGECCO2016}
\bibfield{author}{\bibinfo{person}{Randal~S. Olson}, \bibinfo{person}{Nathan
  Bartley}, \bibinfo{person}{Ryan~J. Urbanowicz}, {and}
  \bibinfo{person}{Jason~H. Moore}.} \bibinfo{year}{2016}\natexlab{}.
\newblock \showarticletitle{Evaluation of a Tree-based Pipeline Optimization
  Tool for Automating Data Science}. In \bibinfo{booktitle}{\emph{Proceedings
  of the Genetic and Evolutionary Computation Conference 2016}} (Denver,
  Colorado, USA) \emph{(\bibinfo{series}{GECCO '16})}.
  \bibinfo{publisher}{ACM}, \bibinfo{address}{New York, NY, USA},
  \bibinfo{pages}{485--492}.
\newblock
\showISBNx{978-1-4503-4206-3}
\urldef\tempurl%
\url{https://doi.org/10.1145/2908812.2908918}
\showDOI{\tempurl}


\bibitem[\protect\citeauthoryear{Pham, Guan, Zoph, Le, and Dean}{Pham
  et~al\mbox{.}}{[n.d.]}]%
        {pham_efficient_2018}
\bibfield{author}{\bibinfo{person}{Hieu Pham}, \bibinfo{person}{Melody~Y.
  Guan}, \bibinfo{person}{Barret Zoph}, \bibinfo{person}{Quoc~V. Le}, {and}
  \bibinfo{person}{Jeff Dean}.} \bibinfo{year}{[n.d.]}\natexlab{}.
\newblock \showarticletitle{Efficient Neural Architecture Search via Parameter
  Sharing}.
\newblock  (\bibinfo{year}{[n.\,d.]}).
\newblock
\showeprint[arxiv]{1802.03268}
\urldef\tempurl%
\url{http://arxiv.org/abs/1802.03268}
\showURL{%
\tempurl}


\bibitem[\protect\citeauthoryear{Ramachandran, Zoph, and Le}{Ramachandran
  et~al\mbox{.}}{2018}]%
        {ramachandran2018searching}
\bibfield{author}{\bibinfo{person}{Prajit Ramachandran},
  \bibinfo{person}{Barret Zoph}, {and} \bibinfo{person}{Quoc~V. Le}.}
  \bibinfo{year}{2018}\natexlab{}.
\newblock \bibinfo{title}{Searching for Activation Functions}.
\newblock
\newblock
\urldef\tempurl%
\url{https://openreview.net/forum?id=SkBYYyZRZ}
\showURL{%
\tempurl}


\bibitem[\protect\citeauthoryear{Real, Aggarwal, Huang, and Le}{Real
  et~al\mbox{.}}{[n.d.]}]%
        {real_regularized_2018}
\bibfield{author}{\bibinfo{person}{Esteban Real}, \bibinfo{person}{Alok
  Aggarwal}, \bibinfo{person}{Yanping Huang}, {and} \bibinfo{person}{Quoc~V.
  Le}.} \bibinfo{year}{[n.d.]}\natexlab{}.
\newblock \showarticletitle{Regularized Evolution for Image Classifier
  Architecture Search}.
\newblock  (\bibinfo{year}{[n.\,d.]}).
\newblock
\showeprint[arxiv]{1802.01548}
\urldef\tempurl%
\url{http://arxiv.org/abs/1802.01548}
\showURL{%
\tempurl}


\bibitem[\protect\citeauthoryear{Salim, Uram, Childers, Balaprakash,
  Vishwanath, and Papka}{Salim et~al\mbox{.}}{[n.d.]}]%
        {salim_balsam_2019}
\bibfield{author}{\bibinfo{person}{Michael~A. Salim},
  \bibinfo{person}{Thomas~D. Uram}, \bibinfo{person}{J.~Taylor Childers},
  \bibinfo{person}{Prasanna Balaprakash}, \bibinfo{person}{Venkatram
  Vishwanath}, {and} \bibinfo{person}{Michael~E. Papka}.}
  \bibinfo{year}{[n.d.]}\natexlab{}.
\newblock \showarticletitle{Balsam: Automated Scheduling and Execution of
  Dynamic, Data-Intensive {HPC} Workflows}.
\newblock  (\bibinfo{year}{[n.\,d.]}).
\newblock
\showeprint[arxiv]{1909.08704}
\urldef\tempurl%
\url{http://arxiv.org/abs/1909.08704}
\showURL{%
\tempurl}


\bibitem[\protect\citeauthoryear{Sergeev and Del~Balso}{Sergeev and
  Del~Balso}{[n.d.]}]%
        {sergeev_horovod_2018}
\bibfield{author}{\bibinfo{person}{Alexander Sergeev} {and}
  \bibinfo{person}{Mike Del~Balso}.} \bibinfo{year}{[n.d.]}\natexlab{}.
\newblock \showarticletitle{Horovod: fast and easy distributed deep learning in
  {TensorFlow}}.
\newblock  (\bibinfo{year}{[n.\,d.]}).
\newblock
\showeprint[arxiv]{1802.05799}
\urldef\tempurl%
\url{http://arxiv.org/abs/1802.05799}
\showURL{%
\tempurl}


\bibitem[\protect\citeauthoryear{Shahriari, Swersky, Wang, Adams, and
  De~Freitas}{Shahriari et~al\mbox{.}}{2015}]%
        {shahriari2015taking}
\bibfield{author}{\bibinfo{person}{Bobak Shahriari}, \bibinfo{person}{Kevin
  Swersky}, \bibinfo{person}{Ziyu Wang}, \bibinfo{person}{Ryan~P Adams}, {and}
  \bibinfo{person}{Nando De~Freitas}.} \bibinfo{year}{2015}\natexlab{}.
\newblock \showarticletitle{Taking the human out of the loop: A review of
  Bayesian optimization}.
\newblock \bibinfo{journal}{\emph{Proc. IEEE}} \bibinfo{volume}{104},
  \bibinfo{number}{1} (\bibinfo{year}{2015}), \bibinfo{pages}{148--175}.
\newblock


\bibitem[\protect\citeauthoryear{Thornton, Hutter, Hoos, and
  Leyton-Brown}{Thornton et~al\mbox{.}}{[n.d.]}]%
        {thornton_auto-weka_2013}
\bibfield{author}{\bibinfo{person}{C. Thornton}, \bibinfo{person}{F. Hutter},
  \bibinfo{person}{H.~H. Hoos}, {and} \bibinfo{person}{K. Leyton-Brown}.}
  \bibinfo{year}{[n.d.]}\natexlab{}.
\newblock \showarticletitle{Auto-{WEKA}: Combined Selection and Hyperparameter
  Optimization of Classification Algorithms}. In
  \bibinfo{booktitle}{\emph{Proc. of {KDD}-2013}} (2013).
  \bibinfo{pages}{847--855}.
\newblock


\bibitem[\protect\citeauthoryear{Wozniak, Jain, Balaprakash, Ozik, Collier,
  Bauer, Xia, Brettin, Stevens, Mohd-Yusof, et~al\mbox{.}}{Wozniak
  et~al\mbox{.}}{2018}]%
        {wozniak2018candle}
\bibfield{author}{\bibinfo{person}{Justin~M Wozniak}, \bibinfo{person}{Rajeev
  Jain}, \bibinfo{person}{Prasanna Balaprakash}, \bibinfo{person}{Jonathan
  Ozik}, \bibinfo{person}{Nicholson~T Collier}, \bibinfo{person}{John Bauer},
  \bibinfo{person}{Fangfang Xia}, \bibinfo{person}{Thomas Brettin},
  \bibinfo{person}{Rick Stevens}, \bibinfo{person}{Jamaludin Mohd-Yusof},
  {et~al\mbox{.}}} \bibinfo{year}{2018}\natexlab{}.
\newblock \showarticletitle{CANDLE/Supervisor: A workflow framework for machine
  learning applied to cancer research}.
\newblock \bibinfo{journal}{\emph{BMC bioinformatics}} \bibinfo{volume}{19},
  \bibinfo{number}{18} (\bibinfo{year}{2018}), \bibinfo{pages}{59--69}.
\newblock


\bibitem[\protect\citeauthoryear{Xia, Shukla, Brettin, Garcia-Cardona, Cohn,
  Allen, Maslov, Holbeck, Doroshow, Evrard, et~al\mbox{.}}{Xia
  et~al\mbox{.}}{2018}]%
        {xia2018predicting}
\bibfield{author}{\bibinfo{person}{Fangfang Xia}, \bibinfo{person}{Maulik
  Shukla}, \bibinfo{person}{Thomas Brettin}, \bibinfo{person}{Cristina
  Garcia-Cardona}, \bibinfo{person}{Judith Cohn}, \bibinfo{person}{Jonathan~E
  Allen}, \bibinfo{person}{Sergei Maslov}, \bibinfo{person}{Susan~L Holbeck},
  \bibinfo{person}{James~H Doroshow}, \bibinfo{person}{Yvonne~A Evrard},
  {et~al\mbox{.}}} \bibinfo{year}{2018}\natexlab{}.
\newblock \showarticletitle{Predicting tumor cell line response to drug pairs
  with deep learning}.
\newblock \bibinfo{journal}{\emph{BMC bioinformatics}} \bibinfo{volume}{19},
  \bibinfo{number}{18} (\bibinfo{year}{2018}), \bibinfo{pages}{486}.
\newblock


\bibitem[\protect\citeauthoryear{Zela, Klein, Falkner, and Hutter}{Zela
  et~al\mbox{.}}{[n.d.]a}]%
        {zela_towards_2018}
\bibfield{author}{\bibinfo{person}{Arber Zela}, \bibinfo{person}{Aaron Klein},
  \bibinfo{person}{Stefan Falkner}, {and} \bibinfo{person}{Frank Hutter}.}
  \bibinfo{year}{[n.d.]}\natexlab{a}.
\newblock \showarticletitle{Towards Automated Deep Learning: Efficient Joint
  Neural Architecture and Hyperparameter Search}.
\newblock  (\bibinfo{year}{[n.\,d.]}).
\newblock
\showeprint[arxiv]{1807.06906}
\urldef\tempurl%
\url{http://arxiv.org/abs/1807.06906}
\showURL{%
\tempurl}


\bibitem[\protect\citeauthoryear{Zela, Siems, and Hutter}{Zela
  et~al\mbox{.}}{[n.d.]b}]%
        {zela_nas-bench-1shot1_2020}
\bibfield{author}{\bibinfo{person}{Arber Zela}, \bibinfo{person}{Julien Siems},
  {and} \bibinfo{person}{Frank Hutter}.} \bibinfo{year}{[n.d.]}\natexlab{b}.
\newblock \showarticletitle{{NAS}-{BENCH}-1SHOT1: {BENCHMARKING} {AND}
  {DISSECTING} {ONE}-{SHOT} {NEURAL} {ARCHITECTURE} {SEARCH}}.
\newblock  (\bibinfo{year}{[n.\,d.]}), \bibinfo{pages}{20}.
\newblock


\bibitem[\protect\citeauthoryear{Zheng, Ji, Wang, Ye, Li, Tian, and Tian}{Zheng
  et~al\mbox{.}}{[n.d.]}]%
        {zheng_rethinking_2020}
\bibfield{author}{\bibinfo{person}{Xiawu Zheng}, \bibinfo{person}{Rongrong Ji},
  \bibinfo{person}{Qiang Wang}, \bibinfo{person}{Qixiang Ye},
  \bibinfo{person}{Zhenguo Li}, \bibinfo{person}{Yonghong Tian}, {and}
  \bibinfo{person}{Qi Tian}.} \bibinfo{year}{[n.d.]}\natexlab{}.
\newblock \showarticletitle{Rethinking Performance Estimation in Neural
  Architecture Search}.
\newblock  (\bibinfo{year}{[n.\,d.]}).
\newblock
\showeprint[arxiv]{2005.09917}
\urldef\tempurl%
\url{http://arxiv.org/abs/2005.09917}
\showURL{%
\tempurl}


\bibitem[\protect\citeauthoryear{Zimmer}{Zimmer}{2020}]%
        {zimmer_2020}
\bibfield{author}{\bibinfo{person}{Lucas Zimmer}.}
  \bibinfo{year}{2020}\natexlab{}.
\newblock \bibinfo{title}{data\_2k.zip}.
\newblock
\newblock
\urldef\tempurl%
\url{https://doi.org/10.6084/m9.figshare.11662428.v1}
\showDOI{\tempurl}


\bibitem[\protect\citeauthoryear{Zimmer, Lindauer, and Hutter}{Zimmer
  et~al\mbox{.}}{[n.d.]}]%
        {zimmer_auto-pytorch_2020}
\bibfield{author}{\bibinfo{person}{Lucas Zimmer}, \bibinfo{person}{Marius
  Lindauer}, {and} \bibinfo{person}{Frank Hutter}.}
  \bibinfo{year}{[n.d.]}\natexlab{}.
\newblock \showarticletitle{Auto-{PyTorch} Tabular: Multi-Fidelity
  {MetaLearning} for Efficient and Robust {AutoDL}}.
\newblock  (\bibinfo{year}{[n.\,d.]}).
\newblock
\showeprint[arxiv]{2006.13799}
\urldef\tempurl%
\url{http://arxiv.org/abs/2006.13799}
\showURL{%
\tempurl}


\bibitem[\protect\citeauthoryear{Zoph and Le}{Zoph and Le}{[n.d.]}]%
        {zoph_neural_2016}
\bibfield{author}{\bibinfo{person}{Barret Zoph} {and} \bibinfo{person}{Quoc~V.
  Le}.} \bibinfo{year}{[n.d.]}\natexlab{}.
\newblock \showarticletitle{Neural Architecture Search with Reinforcement
  Learning}.
\newblock  (\bibinfo{year}{[n.\,d.]}).
\newblock
\showeprint[arxiv]{1611.01578}
\urldef\tempurl%
\url{http://arxiv.org/abs/1611.01578}
\showURL{%
\tempurl}


\bibitem[\protect\citeauthoryear{Zoph, Vasudevan, Shlens, and Le}{Zoph
  et~al\mbox{.}}{[n.d.]}]%
        {zoph_learning_2017}
\bibfield{author}{\bibinfo{person}{Barret Zoph}, \bibinfo{person}{Vijay
  Vasudevan}, \bibinfo{person}{Jonathon Shlens}, {and} \bibinfo{person}{Quoc~V.
  Le}.} \bibinfo{year}{[n.d.]}\natexlab{}.
\newblock \showarticletitle{Learning Transferable Architectures for Scalable
  Image Recognition}.
\newblock  (\bibinfo{year}{[n.\,d.]}).
\newblock
\showeprint[arxiv]{1707.07012}
\urldef\tempurl%
\url{http://arxiv.org/abs/1707.07012}
\showURL{%
\tempurl}


\end{thebibliography}

\begin{center}
    \framebox{\parbox{2.5in}{
    The submitted manuscript has been created by UChicago Argonne, LLC, Operator of Argonne National Laboratory (``Argonne''). Argonne, a U.S. Department of Energy Office of Science laboratory, is operated under Contract No. DE-AC02-06CH11357. The U.S. Government retains for itself, and others acting on its behalf, a paid-up nonexclusive, irrevocable worldwide license in said article to reproduce, prepare derivative works, distribute copies to the public, and perform publicly and display publicly, by or on behalf of the Government. The Department of Energy will provide public access to these results of federally sponsored research in accordance with the DOE Public Access Plan. \url{http://energy.gov/downloads/doe-public-access-plan}}}
    \normalsize
\end{center}

\appendix

\section{Appendix}

We conducted additional set of experiments on ALCF's Theta supercomputer, a CPU-based system using 4 tabular datasets from the OpenML \cite{feurer-arxiv19a} benchmark. These experiments were based on an implementation of data parallelism using MPI, Horovod, and the warmup learning strategy. 

\subsection{Implementation details} 

Fig.~\ref{fig:deephyper-arch} shows a high-level overview of the  implementation in CPU-based system. Algorithm~\ref{alg:AgEBO} ran on a single process $\mathcal{P}$. DeepHyper leveraged the Balsam workflow system~\cite{salim_balsam_2019} to schedule the evaluation of architectures concurrently. Specifically, the submit\_evaluation interface of AgEBO calls the Balsam workflow system, which is responsible for running the architecture training on $W$ workers (via mpirun), collecting the validation accuracy values, and returning the results through a get\_finished\_evaluations interface. We allocated  one compute node for the search. 
%For the BO module implementation, we use the scikit-optimize package \cite{tim_head_2018_1207017} and its ask-and-tell interface. The random forest method is used as the model $M$ within BO.     
We used the Horovod library \cite{sergeev_horovod_2018} for the distributed data-parallel training implementation within AgEBO. 
The AgEBO-Tabular code is open-source and accessible on the DeepHyper GitHub repo.\footnote{\href{https://github.com/deephyper/NASBigData}{https://github.com/deephyper/NASBigData}} 

\begin{figure}[!ht]
    \centering
    \includegraphics[width=0.7\linewidth]{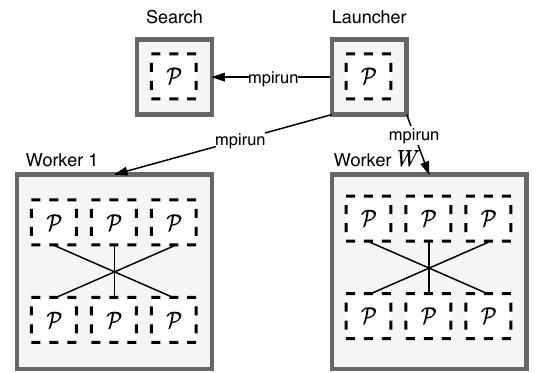}
    \caption{Overview of AgEBO implementation. The AgEBO search runs on a single process and uses the Balsam workflow system to run the architecture evaluation on $W$ workers using the mpirun interface.}
    \label{fig:deephyper-arch}
\end{figure}

\subsection{Experiments}

We used four tabular data sets from the OpenML \cite{feurer-arxiv19a} benchmark. The selection was motivated by a tabular data benchmark study using AutoGluon \cite{erickson_autogluon-tabular_2020}, a recently proposed state-of-the-art AutoML method for tabular data. Among all the data sets benchmarked with AutoGluon, we selected the following four largest data sets having the largest number of data points:
\begin{enumerate}
    \item Covertype \cite{ucikddarchive}: It contains 581,012 data points, 54 input features, and 7 classes.  The task is to predict the forest cover type given cartographic variable input data. 
    \item Airlines \cite{airlinesdataset}: It contains 539,383 data points, 8 input features, and 2 classes. The task is to develop a model to indicate whether a given flight will be delayed or not given input data of the scheduled departure. 
    \item Albert \cite{automlchallenges}: It contains 425,240 data points, 79 input features, and 2 classes from the AutoML Challenge series (2015--2018).
    \item Dionis \cite{automlchallenges}: It contains 416,188 data points, 61 input features, and 355 classes from the AutoML Challenge series (2015--2018).
\end{enumerate}
For each data set, we grouped the data for training, validation, and testing as in the Auto-PyTorch benchmark study. Specifically, we used 42\% for training, 25\% for validation, and 33\% for testing. In all the AutoML methods, we used the training and validation data set within AgEBO-Tablular. The selected best model was evaluated on the testing data.

Experiments were run on the Theta supercomputer at the Argonne Leadership Computing Facility (ALCF). Theta is a Cray XC40 11.69-petaflops system composed of 4,392 nodes with Intel Knights Landing CPUs of 64 cores each equipped of 192 GB of DDR4 memory. Since the data set that we consider fits in a single-node memory, we did not utilize multinode data-parallel training. Instead, the data-parallel training within AgEBO was limited to single node; however, it uses multiple processes within the single node to accelerate training. 

The number of threads per process within the single node, $tpr$, is set to the ratio of the number of threads per node, $tpn$, and the number of process per node, $rpn$. The threading is configured based on guidelines provided by the ALCF, which is based on TensorFlow documentation: intrathreads = \texttt{OMP\_NUM\_THREADS} = $tpr$; interthreads = 2; CPU affinity = depth (equivalent to: \texttt{KMP\_AFFINITY} = ``granularity=fine,verbose,compact,1,0''); \texttt{KMP\_BLOCK\_TIME} = 0.

By default, the NAS experiments were run for a wall time of 3 hours on 129 nodes of Theta. One node was reserved for the search, and 128 nodes were used as workers to train and validate the models within AgEBO.

AgE was used as the baseline. The optimizer was set to Adam~\cite{kingma2014adam}, and each model was evaluated for 20 epochs of training. A gradual warmup strategy \cite{goyal_accurate_2018} was employed for the first 5 epochs. A callback was used to automatically reduce the learning rate on a plateau with a patience of 5 epochs. The objective in the AutoML methods is to maximize the validation accuracy. For the search, the population ($P$) and sample sizes ($S$) were set to $100$ and $10$, respectively. The batch size and learning rate were set to 256 and 0.01, respectively. AgEBO variants adopt the same training strategy as AgE uses. The difference between AgEBO variants and AgE is that the values of the batch size, learning rate, and  number of processes for data-parallel training can be tuned concurrently along with the architecture search.

The range of the hyperparameters for the data-parallel training was set as follows: batch size ($bs_1$) $\in$ [32, 64, 128, 256, 512, 1024]; learning rate ($lr_1$) $\in$ (0.001, 0.1), which are sampled in a log-uniform scale within BO; and  number of processes  ($n$) $\in$ [1,2,4,8].

\subsubsection{Impact of static data-parallel training on AgE}

We show that the accuracy of the architectures discovered by the AgE method with data-parallel training deteriorates significantly without tuning the learning rate, batch size, and  number of processes. 

We evaluated AgE with data-parallel training without BO but varied the number of processes. We used the default learning rate and batch size for $n=1$. The learning rate and batch size for different numbers of processes were scaled by using the linear scaling rule. We ran the experiments on the Covertype data set. 

\begin{figure}[!ht]
    \centering
    \includegraphics[width=0.8\linewidth]{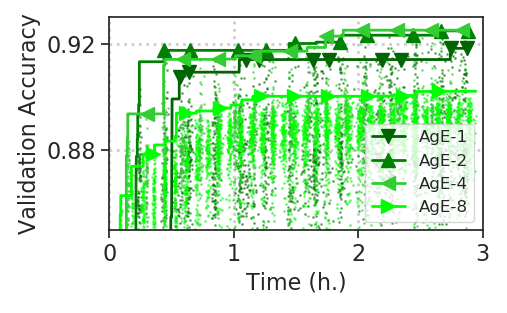}
    \caption{Search trajectory of AgE with different numbers of processes for data-parallel training on the Covertype data set. The thick lines denote the best validation accuracy over time for each method so far. The dots denote the validation accuracy of each architecture found during the search.}
    \label{fig:best-objective-since-start-agebo1}
\end{figure}

The results are shown in Figure \ref{fig:best-objective-since-start-agebo1} and Table \ref{tab:number-evaluated-architectures}, where AgE-$n$ refers to AgE with $n$ processes for data-parallel training. From the results we  observe that increasing the number of ranks from 1 to 4 per evaluation increases the accuracy. This increase can be attributed to the reduced training time for architecture evaluation, which increases the number of evaluated architectures from 632 to 2,421. Nevertheless, for AgE-$8$, we  observe that the accuracy significantly decreases despite the large number (4,221) of evaluated architectures. The poor accuracy of AgE-8 can be attributed to the scaled learning rate and batch size values for 8 processes and/or the possibility that 8 is not the right value for achieving reduction in training time without losing accuracy.

\begin{table}[!ht]
\centering
\resizebox{\linewidth}{!}{%
\begin{tabular}{c|c|c|c|c}
                                                                                     & \textbf{AgE-1} & \textbf{AgE-2} & \textbf{AgE-4} & \textbf{AgE-8} \\ \hline
\textbf{\begin{tabular}[c]{@{}c@{}}Number of\\ architectures\end{tabular}} & 632            & 1764           & 2421           & 4221 \\ \hline
\textbf{Training time (min.)} & $26.54 \pm 7.68$ & $8.97 \pm 0.76$ & $5.38 \pm 0.4$ & $3.19 \pm 0.29$ \\ \hline
\textbf{Validation accuracy} & 0.918 & 0.925 & 0.925 & 0.902  \\ \hline
\end{tabular}}
\caption{
Results for static data-parallel training in AgE.}
\label{tab:number-evaluated-architectures}
\end{table}

\begin{figure}[!ht]
    \centering
    \includegraphics[width=0.8\linewidth]{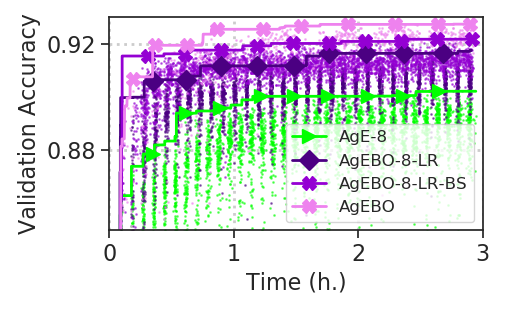}
    \caption{Search trajectory of AgEBO variants and AgE-8 on the Covertype data set. See Fig. \ref{fig:best-objective-since-start-agebo1} caption for the notations used (LR -- learning rate, BS -- batch size).}
    \label{fig:best-objective-since-start-agebo2}
\end{figure}

\subsubsection{Impact of autotuned data-parallel training within AgEBO}

Here we show that tuning the   learning rate, batch size, and  number of processes  through BO improves both the accuracy and time to solution.

To analyze the effectiveness of BO within AgEBO, we compared it with two of its variants. AgEBO-8-LR and AgEBO-8-LR-BS. In the former, only the learning rate was tuned by setting the batch size and the number of processes for the data-parallel training to the default batch size and 8, respectively. In the latter, the batch size and learning rate were tuned by setting the number of processes to 8. As a baseline, we used AgE-8. The experiments were run on the Covertype data set. 

The results are shown in Figure \ref{fig:best-objective-since-start-agebo2}. We   observe that the  AgEBO variants outperform AgE-8 with respect to both accuracy and the time to reach that accuracy. 
The comparison between AgEBO-8-LR and AgE-8 shows that tuning the values of the learning rate leads to significant improvement with respect to both accuracy and time to solution. Similarly, AgEBO-8-LR-BS achieves a higher accuracy value than that of AgEBO-8-LR within a shorter time. However, AgEBO, which tunes all three hyperparameters, outperforms AgEBO-8-LR-BS. An exception is in the initial phases of the search (first 30 minutes), which is due to the initial rank exploration of AgEBO and its impact on the training time. Specifically, this can be attributed to the exploration of different parallelism settings during that phase, which increases the  evaluation time of the architectures.

\begin{figure}[!ht]
    \centering
    \includegraphics[width=0.8\linewidth]{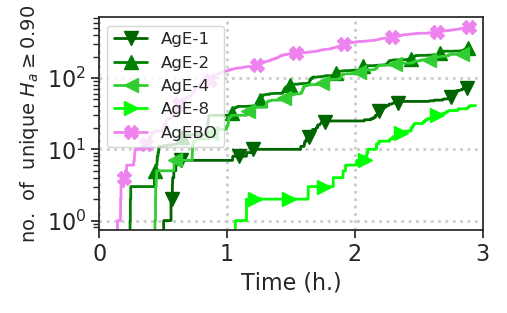}
    \caption{Number of unique high-performing models obtained by AgEBO and AgE-$n$ variants on the Covertype data set.}
    \label{fig:count-of-high-performing-architectures}
\end{figure}

To ensure that the observed superior accuracy of AgEBO is not by chance, we analyzed the number of unique architectures found over time that have a validation accuracy higher than 0.90 for AgE-$n$ variants and AgEBO. The threshold of 0.90 is computed by taking the minimum of 0.99 quantiles of validation accuracy for each variant. The results are shown in Figure \ref{fig:count-of-high-performing-architectures}. We  observe that AgEBO obtains a larger number of high-performing architectures than that of AgE-$n$ variants. Moreover, despite given the same number of nodes, AgEBO is twice as fast as AgE-$n$ variants in reaching the same number of high-performing architectures. Specifically, AgE-$4$ and AgE-$8$ obtain $10^2$ high-performing architectures in 180 minutes whereas AgEBO obtains the same number within 90 minutes. 

\subsubsection{Comparison with AutoGluon and Auto-PyTorch}

Here we show that the prediction accuracy of our method is better than or comparable to that of the two state-of-the-art AutoML software tools AutoGluon \cite{erickson_autogluon-tabular_2020} and Auto-PyTorch \cite{zimmer_auto-pytorch_2020} while reducing the inference time of final models. % INTRO-EXP

The two methods rely on ensemble approaches to boost their prediction accuracy values. AutoGluon combines different supervised learning models such as neural networks, LightGBM, CatBoost, random forest, extra trees, and K-nearest neighbors, the hyperparameters of which are automatically tuned.
%\cite{ke_lightgbm_2017}, CatBoost \cite{dorogush_catboost_nodate}, RandomForest \cite{10.1023/A:1010933404324}, ExtraTrees \cite{10.1007/s10994-006-6226-1} and K-Nearest Neighbours \cite{549118}, 
%Each of these models have different hyperparameters, which are automatically tuned. 
On the other hand, Auto-PyTorch adopts only neural network models but uses an ensemble strategy to improve the accuracy. We compared AgEBO with AutoGluon and Auto-PyTorch on all four data sets. We used AgE-1 as a baseline. 

AutoGluon was run on a single node with a time limit of 4 hours for the call to the fit method to compensate for  possible issues with the time estimation performed by the software. The \texttt{hyperparameter\_tune=True} and \texttt{auto\_stack=True} were set to maximize  the accuracy as much as possible. The test accuracy was computed separately by reloading the saved models. 
Table \ref{tab:benchmark} shows the accuracy values of the best models and the corresponding inference time of AgEBO and AutoGluon. We  observe that the test accuracy values of AgEBO and AutoGluon are comparable on all four data sets. However, the key advantage stems from the inference time with the trained model. Given that AgEBO generates a single neural network model, the inference time is between 2.7 and 4.3 seconds. On the other hand, AutoGluon relies on stacking a number of models,  resulting in an inference time of about 7 minutes.

For Auto-PyTorch, since we cannot install the software in our ALCF Theta software stack because of software dependency issues, we used the results from the LCBench data base \cite{zimmer_2020}, which stores the results of experimental runs of the four data sets.  We note, however,  that although we used the same proportion of the training, validation, and testing split, the exact data splits were not used, the details of which are not available. Moreover, we did not compare against test accuracy from the ensemble strategy from Auto-PyTorch because we cannot retrieve ensemble strategy results from the LCBench database. Therefore, we focus on comparison with validation accuracy values.

\begin{figure*}[!ht]
    \centering

    \begin{subfigure}{0.24\linewidth}
        \centering
        \includegraphics[width=0.99\textwidth]{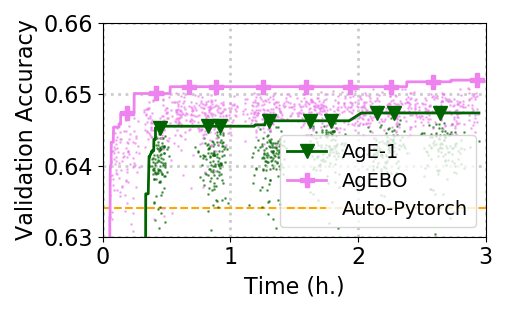}
        \caption{Airlines}
        \label{fig:airlines-best-objective-since-start}
    \end{subfigure}
    \begin{subfigure}{0.24\linewidth}
        \centering
        \includegraphics[width=0.99\textwidth]{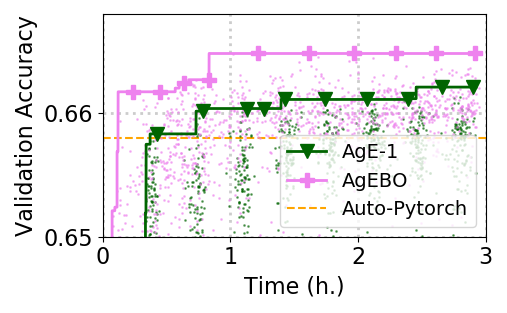}
        \caption{Albert}
        \label{fig:albert-best-objective-since-start}
    \end{subfigure}
    \begin{subfigure}{0.24\linewidth}
        \centering
        \includegraphics[width=0.99\textwidth]{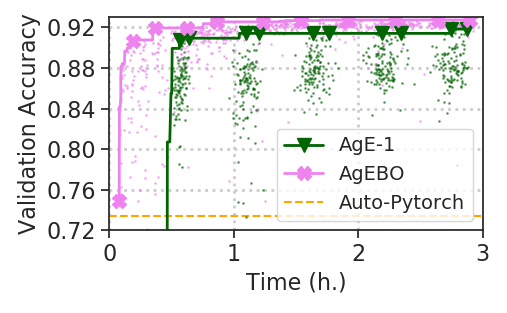}
        \caption{Covertype}
        \label{fig:covertype-best-objective-since-start}
    \end{subfigure}
    \begin{subfigure}{0.24\linewidth}
        \centering
        \includegraphics[width=0.99\textwidth]{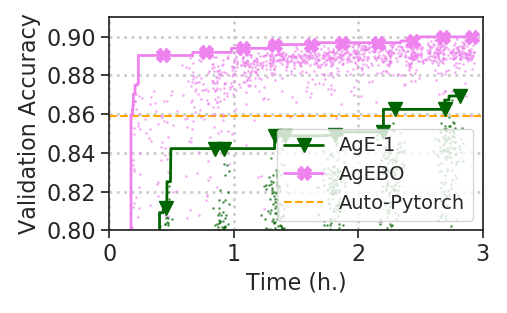}
        \caption{Dionis}
        \label{fig:dionis-best-objective-since-start}
    \end{subfigure}
    \caption{
    Search trajectory of AgE-1, AgEBO, and Auto-Pytorch on the four data sets.
    A horizontal dotted line shows the validation accuracy at the 20\ts{th} epoch of the model with the best validation accuracy found by Auto-PyTorch. See Fig. \ref{fig:best-objective-since-start-agebo1} caption for the notations used.
    }
    \label{fig:all-experiments-time-to-solution-results}
\end{figure*}

Figure \ref{fig:all-experiments-time-to-solution-results} shows the comparison between the best validation accuracy values  found by AgEBO and Auto-PyTorch. 
We observe that AgEBO achieves validation accuracy values that are higher than those of Auto-PyTorch within 30 minutes of search time. The  differences in the accuracy values can be explained by two factors. First, Auto-PyTorch is not designed to generate a single neural network model but to generate multiple models and combine them using an ensemble strategy to have good accuracy. Second, the architecture space of Auto-PyTorch is restricted to a smaller number of trainable parameters and smaller number of layers. 

%However, it was shown in Auto-PyTorch work using a nonparameteric sensitivity approach  that the number of layers has higher importance on the accuracy.

%An additional remark as to explain the good selection of our model based on a selection at the 20\ts{th} epoch is brought by the LCBench study \cite{zimmer_auto-pytorch_2020} which shows good spearman rank correlation for Airlines, Albert, Covertype and Dionis between budgets of 25 and 50 epochs which explain the good performance of our final models during a longer training (i.e., more epochs) based on a selection with a 20 epochs budget. The authors of Auto-PyTorch also explained that the number of layers has a high feature importance in the model performance which confirms our choice of allowing a maximum of 10 layers.

The comparison between AgE-1 and AgEBO in Figure \ref{fig:all-experiments-time-to-solution-results} summarizes the benefits of autotuned data-parallel training. For the Airlines data set, the maximal accuracy found with AgE-1 is 0.647 at 121 minutes, whereas  AgEBO  finds a greater accuracy after 14 minutes and reaches its maximal accuracy of 0.652 after 163 minutes. For the Albert data set, the maximal accuracy found with AgE-1 is 0.662 at 147 minutes, whereas  AgEBO achieves a higher accuracy after 36 minutes and  reaches its maximal accuracy of 0.665 after 49 minutes. For Covertype, the maximal accuracy found with AgE-1 is 0.918 at 164 minutes, whereas  AgEBO achieves a greater accuracy after 20 minutes and reaches its maximal accuracy of 0.927 after 165 minutes. For the Dionis data set, the maximal accuracy found with AgE-1 is 0.869 at 163 minutes, whereas  AgEBO achieves a greater accuracy after 11 minutes and reaches its maximal accuracy of 0.900 after 147 minutes. In summary, AgEBO outperforms the AgE-1 with respect to both accuracy values and time to reach those accuracy values. 

\begin{table}[!ht]
\centering
\resizebox{\linewidth}{!}{%
\begin{tabular}{c|c|c|c|c|}
\cline{2-5}
\textbf{}                              & \multicolumn{2}{c|}{\textbf{\begin{tabular}[c]{@{}c@{}}AgEBO \end{tabular}}} & \multicolumn{2}{c|}{\textbf{AutoGluon}}                                       \\ \hline
\multicolumn{1}{|c|}{\textbf{data set}} & \multicolumn{1}{l|}{\textbf{Test}}      & \multicolumn{1}{l|}{\textbf{Inference}}      & \multicolumn{1}{l|}{\textbf{Test}} & \multicolumn{1}{l|}{\textbf{Inference }} \\ %\hline
\multicolumn{1}{|c|}{} & \multicolumn{1}{l|}{\textbf{Accuracy}}      & \multicolumn{1}{l|}{\textbf{Time (s)}}      & \multicolumn{1}{l|}{\textbf{Accuracy}} & \multicolumn{1}{l|}{\textbf{Time (s)}} \\ \hline
\multicolumn{1}{|c|}{Airlines}         & \textbf{0.652 $\pm$ 0.002}                              & 3.1                                       & 0.641                                  & 1124.9                               \\
\multicolumn{1}{|c|}{Albert}           & 0.661 $\pm$ 0.001                                       & 2.7                                       & \textbf{0.688}                         & 409.3                                \\
\multicolumn{1}{|c|}{Covertype}        & \textbf{0.963 $\pm$ 0.001}                              & 4.3                                       & 0.961                                  & 906.6                                \\
\multicolumn{1}{|c|}{Dionis}           & \textbf{0.915 $\pm$ 0.0005}                              & 3.2                                       & 0.907                                  & 1900.5                               \\ \hline
\end{tabular}}
\caption{Test accuracy values and inference times obtained by AgEBO and AutoGluon on the four data sets.
}
\label{tab:benchmark}
\end{table}

Across all four data sets we observed that the node utilization of AgEBO is similar to that of AgE---both reach an average value of $\approx$94\%. This can be attributed to the effectiveness of the asynchronous BO that generates hyperparameter configurations with minimal overhead, which are combined with architecture decision variable values and sent for evaluation with minimal delay.

Table \ref{tab:top-5-configurations} shows the best hyperparameters obtained  by AgEBO for the top 5 best-performing models on the four data sets. Note that AgEBO finds different hyperparameter configurations for different data sets to accelerate data-parallel training. Within the same data set, the hyperparameter configurations obtained for the best models are similar. These results demonstrate the need for data-set-specific hyperparameter tuning for data-parallel training, which is enabled by AgEBO.

\begin{figure}[!ht]
    \centering
    \includegraphics[width=1.\linewidth]{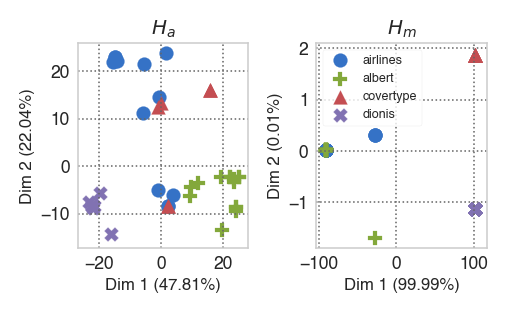}
    \caption{Principal component analysis projection of top 1\% configurations of architecture decision variables ($H_a$) and data-parallel training hyperparameters ($H_m$). The \% on each axis shows the conserved variance (more than 80\%) in two-dimensional projections.}
    \label{fig:pca_ha_hm}
\end{figure}

We visualized the top 1\% configurations based on the validation accuracy values obtained on all four data sets using principal component analysis. This is done  by projecting the 37 architecture decisions and 3 hyperparameters of the top 1\% configurations into two dimensions, respectively. The results are shown in Figure \ref{fig:pca_ha_hm}. From the results we can see a similar pattern. Each data set requires different values for architecture decision variables and data-parallel training hyperparameters.

\begin{table}[!ht]
\centering
\resizebox{0.8\linewidth}{!}{%
\begin{tabular}{c|c|c|c|c}
                                    & \textbf{\begin{tabular}[c]{@{}c@{}}batch\\ size\end{tabular}} & \textbf{\begin{tabular}[c]{@{}c@{}}learning \\ rate\end{tabular}} & \textbf{\begin{tabular}[c]{@{}c@{}}no.\ of  \\ processes\end{tabular}} & \textbf{\begin{tabular}[c]{@{}c@{}}validation\\ accuracy\end{tabular}} \\ \hline
\multirow{5}{*}{\textbf{Airlines}}  & 64.0                                                          & 0.001474                                                          & 2.0                                                                & 0.652008                                                               \\
                                    & 64.0                                                          & 0.001250                                                          & 2.0                                                                & 0.651774                                                               \\
                                    & 128.0                                                         & 0.001541                                                          & 2.0                                                                & 0.651086                                                               \\
                                    & 128.0                                                         & 0.001742                                                          & 2.0                                                                & 0.651086                                                               \\
                                    & 64.0                                                          & 0.001538                                                          & 2.0                                                                & 0.65090                                                                \\ \hline
\multirow{5}{*}{\textbf{Albert}}    & 128.0                                                         & 0.005726                                                          & 4.0                                                                & 0.664827                                                               \\
                                    & 64.0                                                          & 0.002226                                                          & 2.0                                                                & 0.664808                                                               \\
                                    & 64.0                                                          & 0.002304                                                          & 2.0                                                                & 0.664552                                                               \\
                                    & 64.0                                                          & 0.002490                                                          & 2.0                                                                & 0.664446                                                               \\
                                    & 64.0                                                          & 0.002154                                                          & 2.0                                                                & 0.664190                                                               \\ \hline
\multirow{5}{*}{\textbf{Covertype}} & 256.0                                                         & 0.001392                                                          & 1.0                                                                & 0.927418                                                               \\
                                    & 256.0                                                         & 0.001371                                                          & 1.0                                                                & 0.927325                                                               \\
                                    & 256.0                                                         & 0.001409                                                          & 1.0                                                                & 0.927317                                                               \\
                                    & 256.0                                                         & 0.001394                                                          & 1.0                                                                & 0.927309                                                               \\
                                    & 256.0                                                         & 0.001394                                                          & 1.0                                                                & 0.927294                                                               \\ \hline
\multirow{5}{*}{\textbf{Dionis}}    & 256.0                                                         & 0.001201                                                          & 4.0                                                                & 0.899902                                                               \\
                                    & 256.0                                                         & 0.001237                                                          & 4.0                                                                & 0.899192                                                               \\
                                    & 256.0                                                         & 0.001211                                                          & 4.0                                                                & 0.898837                                                               \\
                                    & 256.0                                                         & 0.001159                                                          & 4.0                                                                & 0.898482                                                               \\
                                    & 256.0                                                         & 0.001159                                                          & 4.0                                                                & 0.898260                                                              
\end{tabular}}
\caption{Data-parallel training hyperparameter values obtained by AgEBO for the top 5 best models on the four data sets.}
\label{tab:top-5-configurations}
\end{table}

\subsubsection{Exploration and exploitation in AgEBO}
Here we study the effect of exploration and exploitation of BO within AgEBO by varying $\kappa$ values. We  show that stronger exploitation is critical for the effectiveness of AgEBO.

The $\kappa$ value in Eq.~\ref{eqn:ucb} controls the trade-off between exploration and exploitation in BO. In addition to the default $\kappa$ value of 0.001, we ran AgEBO with two values: \{1.96, 19.6\}. Note that $1.96$ is the typical $\kappa$ value in Scikit-Optimize, which provides a balance between exploration and exploitation. The value of 19.6 is selected to enforce large exploration. We ran the experiments on the Covertype and the Dionis data sets.

\begin{figure}[!h]
    \centering
    \begin{subfigure}{0.49\linewidth}
        \centering
        \includegraphics[width=1.\textwidth]{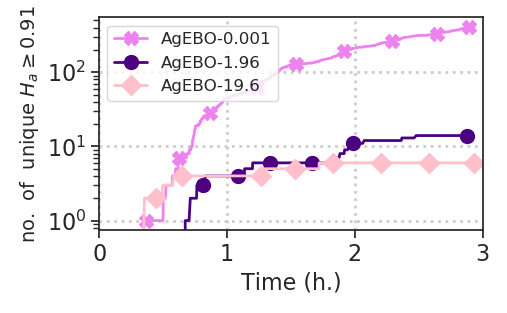}
        \caption{Covertype}
        \label{fig:kappa-count-covertype}
    \end{subfigure}
    \begin{subfigure}{0.49\linewidth}
        \centering
        \includegraphics[width=1.\textwidth]{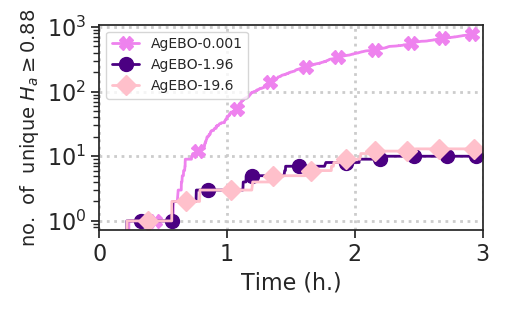}
        \caption{Dionis}
        \label{fig:kappa-count-dionis}
    \end{subfigure}

    \caption{Number of unique high-performing architectures discovered by AgEBO over time with different $\kappa$ values.
    }
    \label{fig:kappa-count-plots}
\end{figure}

Figure \ref{fig:kappa-count-plots} shows the number of high-performing architectures found by AgEBO for three different $\kappa$ values. The threshold was computed by computing 99\% quantiles of the validation accuracy values for the three variants and taking the smallest value. We   observe that for both  data sets, AgEBO with the default $\kappa$ value of 0.001 (stronger  exploitation) completely outperforms those with 1.96 (balance between exploration and exploitation) and 19.6 (stronger exploration) with respect to the number of high-performing architectures (between one and two orders of magnitude) and time needed to reach the number of the other two variants (between 2x and 3x faster).
The exploration of hyperparameter values in AgEBO with a  $\kappa$ value of 0.001 happens only in the random initialization phase. During the iterative phase, given the stronger exploitation setting, hyperparameter configurations are generated close the best ones found so far in the search. On the other hand, there is a significant degree of exploration with $\kappa$ values of 1.96 and 19.6. This results in increased data-parallel training time, which eventually reduces the generation of number of high-performing architectures.

\end{document}